%% file: acl_latex.tex
\pdfoutput=1

\documentclass[11pt]{article}

\usepackage[]{acl}

\usepackage{times}
\usepackage{latexsym}
\usepackage[T1]{fontenc}

\usepackage[utf8]{inputenc}
\definecolor{hidden-draw}{RGB}{205, 44, 36}
\definecolor{hidden-blue}{RGB}{194,232,247}
\definecolor{hidden-orange}{RGB}{243,202,120}
\definecolor{hidden-yellow}{RGB}{255,229,204}
\definecolor{hidden-red}{RGB}{255,204,204}
\usepackage{microtype}
\usepackage{xcolor,colortbl}
\usepackage{marvosym}
\usepackage{bm,amsmath}
\usepackage{multirow}
\usepackage{makecell}
\usepackage{xspace,mfirstuc,tabulary}
\usepackage{booktabs}
\usepackage{hyperref}
\usepackage{color}
\usepackage{tikz}
\usepackage{graphicx}
\usepackage{inconsolata}
\usepackage{pifont}
\usepackage[edges]{forest}
\definecolor{hidden-draw}{RGB}{20,68,106}
\definecolor{hidden-pink}{RGB}{255,245,247}

\title{\includegraphics[width=48pt,height=24pt]{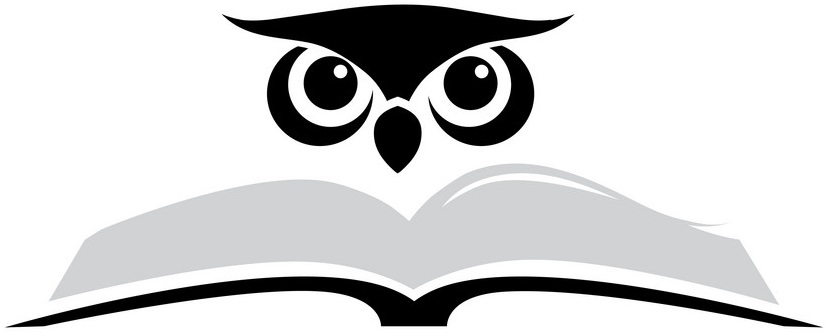}~Trends in Integration of Knowledge and Large Language Models: \\ A Survey and Taxonomy of Methods, Benchmarks, and Applications}

\author{
    Zhangyin Feng$^{1}$\footnotemark[1],
    Weitao Ma$^{1}$\thanks{$\quad$ means Equal Contribution, \Letter~ means Corresponding Author},
    Weijiang Yu$^{2\textsuperscript{\Letter}}$,
    Lei Huang$^{1}$,
    Haotian Wang$^{1}$, \\
    \textbf{Qianglong Chen$^{2}$,
    Weihua Peng$^{2}$,
    Xiaocheng Feng$^{1\textsuperscript{\Letter}}$,
    Bing Qin$^{1}$,
    Ting Liu$^{1}$ }\\
    $^{1}$Harbin Institute of Technology, Harbin, China \\
    $^{2}$Huawei Inc., Shenzhen, China \\
    \texttt{ \{zyfeng,wtma,lhuang,xcfeng,qinb,tliu\}@ir.hit.edu.cn} \\
    \texttt{\{weijiangyu8,wanght1998,chenqianglong.ai,pengwh.hit\}@gmail.com}
}

\begin{document}
\maketitle
\begin{abstract}
Large language models (LLMs) exhibit superior performance on various natural language tasks, but they are susceptible to issues stemming from outdated data and domain-specific limitations.
In order to address these challenges, researchers have pursued two primary strategies, knowledge editing and retrieval augmentation, to enhance LLMs by incorporating external information from different aspects.
Nevertheless, there is still a notable absence of a comprehensive survey. 
In this paper, we propose a review to discuss the trends in integration of knowledge and large language models, including taxonomy of methods, benchmarks, and applications.
In addition, we conduct an in-depth analysis of different methods and point out potential research directions in the future.
We hope this survey offers the community quick access and a comprehensive overview of this research area, with the intention of inspiring future research endeavors.

\end{abstract}

\section{Introduction}
Large language models (LLMs) have demonstrated an impressive ability to encode real-world knowledge in their parameters and a remarkable capacity for solving various natural language processing tasks \cite{brown2020language, hoffmann2022training, zeng2022glm, chowdhery2022palm, touvron2023llama, zhao2023survey}.
However, they still suffer from serious challenges in knowledge-intensive tasks~\cite{petroni-etal-2021-kilt}, which require a substantial volume of real-world knowledge.

Recent works show that LLMs struggle to learn long-tail knowledge~\cite{kandpal2023large, mallen-etal-2023-trust}, are not able to update their parameters in time to capture the changing world \cite{de-cao-etal-2021-editing, kasai2022realtime} (i.e., the parameters of ChatGPT \footnote{\url{https://chat.openai.com}} only contain information before September 2021, and are completely unaware of the latest world knowledge.), and suffer from hallucinations~\cite{zhang2023sirens, rawte2023survey, huang2023survey}.
To alleviate these problems, there has been growing interest in integrating knowledge and large language models through knowledge editing or retrieval augmentation.
Knowledge editing~\cite{de-cao-etal-2021-editing, sinitsin2020editable} aims to modify obsolete knowledge in LLMs using an efficient method that only updates partial model parameters.
Retrieval augmentation~\cite{mallen-etal-2023-trust, shi2023replug, trivedi2023interleaving} employs an off-the-shelf retrieval model to fetch relevant documents from an external corpus to aid large language models and maintains their parameters unchanged.
Numerous works have been proposed to integrate knowledge and large language models, focusing on the aforementioned two aspects. Nevertheless, these endeavors remain rather fragmented, lacking a comprehensive and systematic review.
\input{taxonomy}

To fill the gap, in this paper, we present a concrete organization of our survey, focusing on knowledge editing and retrieval augmentation, as depicted in Figure~\ref{fig:survey}.
We begin by systematically introducing the knowledge editing methods according to the processed structure of the model (\S \ref{know_edit}), including input editing (\S \ref{input_edit}), model editing (\S \ref{model_edit}) and assess knowledge editing (\S \ref{knowledge_edit_assess}) which covers representative methods and general benchmarks.
Furthermore, we provide a detailed discussion of retrieval augmentation (\S \ref{sec:retrieval_aug}), including retrieval judgement (\S \ref{sec:retrieval_judge}), document retrieval (\S \ref{sec:retrieval_method}), document utilization (\S \ref{sec:document_util}), knowledge conflict (\S \ref{sec:knowledge_conflict}) and benchmark (\S \ref{sec:benchmark}).
Then, we summarize some cutting-edge applications of integration of knowledge and large language models (\S \ref{application}), such as New Bing~\footnote{\url{https://www.bing.com/new}}. 
Finally, to stimulate further research in this field, we offer insights into prospective directions for future investigations (\S \ref{future work}). 

\paragraph{Related Work}
Some previous reviews also discussed the interaction between language models and external knowledge. \citet{hu2023survey} deliberates on the various forms of knowledge and methods employed to augment language models in prior research, which only focuses on pre-trained language models with smaller parameter sizes. \citet{yao2023editing} specializes in a comprehensive and empirical discussion on existing knowledge editing methods which does not involve other retrieval-related knowledge application methods. \citet{zhang2023large} offers a broad and comprehensive overview of the strategies for adapting LLMs to dynamically updated world knowledge, while we provide a deep and detailed analysis of current methods and benchmarks. Other existing knowledge-enhanced LMs surveys \cite{yin2022survey,Yu_2022,yang2023chatgpt} center on pre-trained language models and frequently employ re-training methods to incorporate additional knowledge that is not suitable for the current LLMs. We will comprehensively introduce the recent advancements in integration of knowledge and large language models, with a specific focus on two effective methods: knowledge editing and retrieval augmentation through providing a detailed analysis and offering insights into future developments.

\section{Knowledge Editing}
\label{know_edit}

\begin{figure*}[h]
\centering
\includegraphics[width=\textwidth]{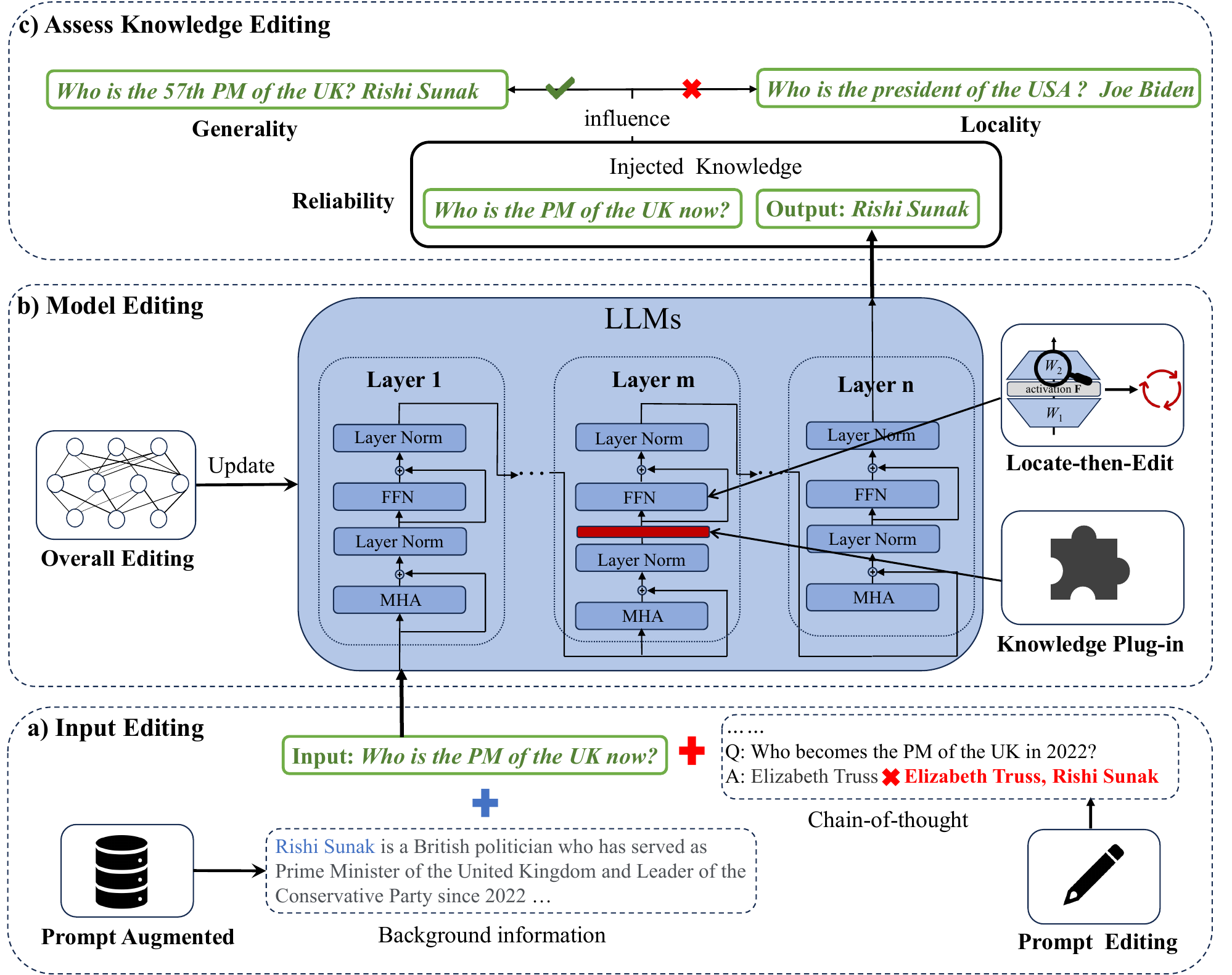}
\caption{An overview of knowledge editing from three perspectives, which encompasses input (input editing), processing (model editing), and output (assess knowledge editing) aspects of the model's processing structure.}
\label{fig:knowledge_edit_picture}
\end{figure*}

Knowledge editing is an emerging method for rectifying inaccuracies and updating outdated information in LLMs through the incorporation of new knowledge. In this section, we delve into the current works about knowledge editing, with a specific focus on the processed structure of LLMs across various methods. As shown in Figure \ref{fig:knowledge_edit_picture},  we organize them into three categories: input editing (\S \ref{input_edit}), model editing (\S \ref{model_edit}), and assess knowledge editing (\S \ref{knowledge_edit_assess}).

\input{knowledge_edit}

\section{Retrieval Augmentation}
\label{sec:retrieval_aug}

As discussed in section \S \ref{know_edit}, knowledge editing ~\cite{de-cao-etal-2021-editing} is a productive approach to updating outdated knowledge by modifying the parameters of a specific part of large language models.
However, knowledge editing faces some other problems. 
Firstly, it is not entirely clear how and where knowledge is stored in large language models. 
Secondly, the mapping relationship between knowledge and parameters is very complicated, and modifying the parameters corresponding to some knowledge may affect other knowledge.
In this section, we introduce retrieval augmentation, an alternative route to integrate knowledge and large language models while keeping the parameters unchanged. 

Unlike knowledge editing, which mainly parameterizes external knowledge to update the large language models, retrieval augmentation utilizes external knowledge in a non-parameterized form in the inference stage.
Retrieval augmentation typically consists of a retriever and a large language model.
Given an input context, the retriever first fetches relevant documents from an external corpus.
Then, we can use relevant documents at different stages to improve the performance of large language models.
In this section, we focus on the following key questions of retrieval augmentation:
\begin{itemize}
    \item When do large language models need to be enhanced by retrieval? (\S \ref{sec:retrieval_judge})
    \item How to retrieve relevant documents? (\S \ref{sec:retrieval_method})
    \item How do large language models utilize retrieved documents? (\S \ref{sec:document_util})
    \item How to resolve knowledge conflicts from different documents? (\S \ref{sec:knowledge_conflict})
\end{itemize}

\subsection{Retrieval Judgement}
\label{sec:retrieval_judge}
A very important problem for retrieval-augmented large language models is to know the knowledge boundaries~\cite{yin-etal-2023-large} of LLMs and determine when to retrieve supplementary knowledge.
The current retrieval judgment methods are mainly divided into two categories: calibration-based judgment and model-based judgment.

\paragraph{Calibration-based.}
\label{rule_base}

A simple and intuitive idea is to set a metric and a threshold.
When the metric is above or below the threshold, we trigger the retriever to fetch relevant documents.
\citet{kandpal2023large} study the relationship between the knowledge memorized by large language models and the information in pre-training datasets.
Their results demonstrate the strong correlational relationship between accuracy and relevant document count for numerous question-answering datasets.
In order to deeply analyze the relationship between the parameterized knowledge of LLMs and the data popularity, \citet{mallen-etal-2023-trust} build an open domain question answering dataset PopQA, which contains entity popularity from Wikipedia.
Then, they devise an adaptive retrieval method to only use retrieval for questions whose popularity is lower than the popularity threshold.
In addition to popularity, \citet{jiang-etal-2021-know} show that LLMs tend to be well-calibrated and low probability or confidence often indicates a lack of relevant knowledge.
\citet{si2022prompting} and \citet{manakul2023selfcheckgpt} utilize the token probability to indicate the uncertainty of their output.
Following this idea, \citet{jiang2023active} propose a confidence-based active retrieval approach, named FLARE. 
If the confidence of each word in the generated sentence is above the threshold, they accept the sentence without retrieving additional information.
Otherwise, they actively trigger retrieval and utilize the retrieved relevant information to regenerate the current sentence.

\begin{figure*}[!h]
  \includegraphics[width=\linewidth]{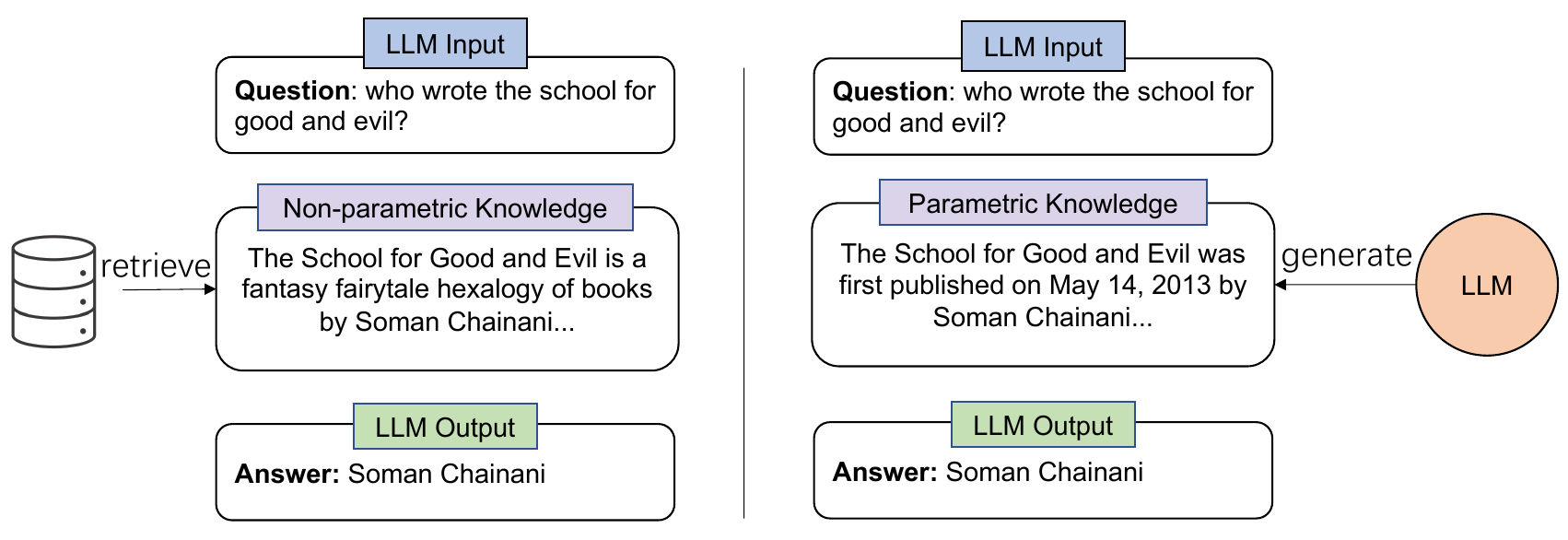}
  \caption{Two kinds of document retrieval methods: the document on the left is fetch from an external corpus with a retriever, and the document on the right is generated by a large language model.}
  \label{fig:retrieval}
\end{figure*}

\paragraph{Model-based.}
\label{model_base}

There are two kinds of model-based judgment methods: \textit{normal setting} and \textit{retrieval setting}.
The \textit{normal setting} is to directly determine whether to trigger retrieval based on the question.
Considering that large language models have very powerful capabilities, some researchers directly employ large language models to determine whether retrieval is needed.
\citet{yin-etal-2023-large} investigate the self-knowledge of LLMs by assessing their ability to identify unanswerable or unknowable questions.
\citet{kadavath2022language} prompt LLMs to predict the probabilities of whether their responses are reliable.
Those unreliable responses indicate that LLMs require additional information to answer the corresponding questions.
\citet{feng2023knowledge} ask LLMs \textit{``Do you need more information? (Yes or No)''} to determine whether external knowledge is needed for the given question through in-context learning.
\citet{ren2023investigating} adopt priori and posteriori judgment instructions to investigate whether LLMs are capable of perceiving their own factual knowledge boundary for both normal setting and retrieval setting.
Priori judgment asks LLMs whether they can provide an answer to the question.
Posteriori judgment asks LLMs to evaluate the correctness of the answer to the question.
They observe that LLMs perceive their factual knowledge boundary inaccurately and have a tendency to be overconfident in normal setting.

The \textit{retrieval setting} is to first retrieve the relevant documents for all questions, and then judge whether the relevant documents can answer the question.
If relevant, large language models utilize the retrieved documents to generate the answer. 
If irrelevant, large language models directly generate the answer.
\citet{Yoran_Wolfson_Ram_Berant_2023} regard the judgment of the relevance of retrieved documents and questions as a natural language inference (NLI) problem \cite{dagan2005pascal, bowman2015large} and use a well-trained BART-Large~\cite{lewis2019bart} NLI model to identify irrelevant retrieved documents.
The retrieved documents serve as the premise, while the question and generated answer are concatenated and serve as hypothesis.
\citet{baek2023knowledgeaugmented}  propose to fine-tune LLMs with instruction data to  identify the relatedness between the input question and the retrieved documents, and ensemble results from various instructions to further improve  accuracy.
\citet{ren2023investigating} find that the accuracy of LLMs' self-assessment improves after incorporating relevant documents and it is effective to dynamically introduce retrieved documents for LLMs.

\noindent 
\textit{Highlight:} Calibration-based methods are simple and effective.
However, these methods are mostly ad-hoc, while the score may not be available for commercial LLMs.
Additionally, it may be challenging to find appropriate thresholds for different data and models due to the sensitivity of thresholds.
Judging whether to trigger retrieval with a well-trained model is a promising direction.
At present, this direction is in an early stage, and there is still a large space for exploration.

\subsection{Document Retrieval}
\label{sec:retrieval_method}

As shown in Figure ~\ref{fig:retrieval}, there are two ways to get the relevant documentation.
One approach is to use a retriever to fetch relevant documents from an external corpus (e.g. Wikipedia).
Another approach is to use a large language model to generate relevant documents.

\paragraph{Retriever-based.}

Given an input context $x$, a retriever aims to retrieve a small set of documents from a corpus $\mathcal{D} = \{d_1...d_m\}$ that are relevant to $x$.
There are different types of searchers, including term-based sparse retriever, embedding-based dense retriever and commercial search engines.
Sparse retriever is usually implemented using TF-IDF or BM25 \cite{robertson2009probabilistic}, which matches keywords efficiently with an inverted index.
However, term-matching methods are sensitive to highly selective keywords
and phrases.
Dense retriever \cite{DBLP:conf/emnlp/KarpukhinOMLWEC20} encodes text into a continuous dense semantic space, where synonyms or paraphrases that consist of completely different tokens may still be mapped to vectors close to each other.
Commercial search engines, such as Google and Baidu, are complex systems that are able to retrieve the latest world knowledge.
These three methods have different advantages and application scenarios, and we will focus on dense retriever next.

Given a collection of text passages, the goal of the dense retriever is to index all the passages in a low-dimensional and continuous space, such that it can retrieve efficiently the top $k$ passages relevant to the input question for the reader at run-time.
Dense retriever uses a dense encoder $E(\cdot)$ which maps any text passage to a $d$-dimensional real-valued vectors.
Specifically, the encoder maps each document $d \in \mathcal{D}$ to an embedding $E(d)$ by taking the mean pooling of the last hidden representation over the tokens in $d$. 
At query time, the same encoder is applied to the input context $q$ to obtain a query embedding $E(q)$. The similarity between the query embedding and the document embedding is computed by their cosine similarity:
\begin{eqnarray}
s(d, q) & = & \cos (E(d), E(q))
\end{eqnarray}
The top-$k$ documents that have the highest similarity scores when compared with the input $q$ are retrieved in this step.

Prior works explore different ways to train the whole retriever-LM system in an end-to-end fashion, using retrieval augmented sequence log-likelihood \cite{lewis2021retrievalaugmented, borgeaud2022improving}, fusion-in-decoder attention distills \cite{izacard2022distilling, izacard2022atlas}, or knowledge graph \cite{ju2022grape}.
This kind of fine-tuning can be expensive when more and more unique demands emerge \cite{maronikolakis-schutze-2021-multidomain}. 
More importantly, many toptier LMs can only be accessed through black-box APIs \cite{ouyang2022training, openai2023gpt4}.
These APIs allow users to submit queries and receive responses but typically do not support fine-tuning.
In contrast to prior work that adapts language models to the retriever, recent work attempts to adapt the retriever to language models.
REPLUG LSR \cite{shi2023replug} further improves the initial retrieval model in REPLUG with supervision signals from a black-box language model, i.e. GPT-3 Curie \cite{brown2020language}. 
AAR \cite{yu-etal-2023-augmentation} proposes to leverage a small source LM to provide LM-preferred signals for the retriever’s training. 
The retriever after training can be directly utilized to assist a large target LM by plugging in the retrieved documents.

Unlike prior studies focusing on adapting either the retriever or the language models,
another research line focuses on bridging the semantic gap between the input text and the knowledge that is really needed to query.
Query2doc~\cite{wang2023query2doc} prompts the LLMs to generate a pseudo-document by employing a few-shot prompting paradigm.
Subsequently, the original query is expanded by incorporating the pseudo-document.
The retriever module uses this new query to retrieve a list of relevant documents.
\citet{ma2023query} introduce a Rewrite-Retrieve-Read framework for retrieval augmentation, which can be further tuned for adapting to LLMs.
They also add a query rewriting step before the retriever. 
Different with Query2doc, they adopt a trainable language model to perform the rewriting step.
The rewriting language model is trained by reinforcement learning to using the LLM performance as a reward.

\begin{figure*}[!h]
  \includegraphics[width=\linewidth]{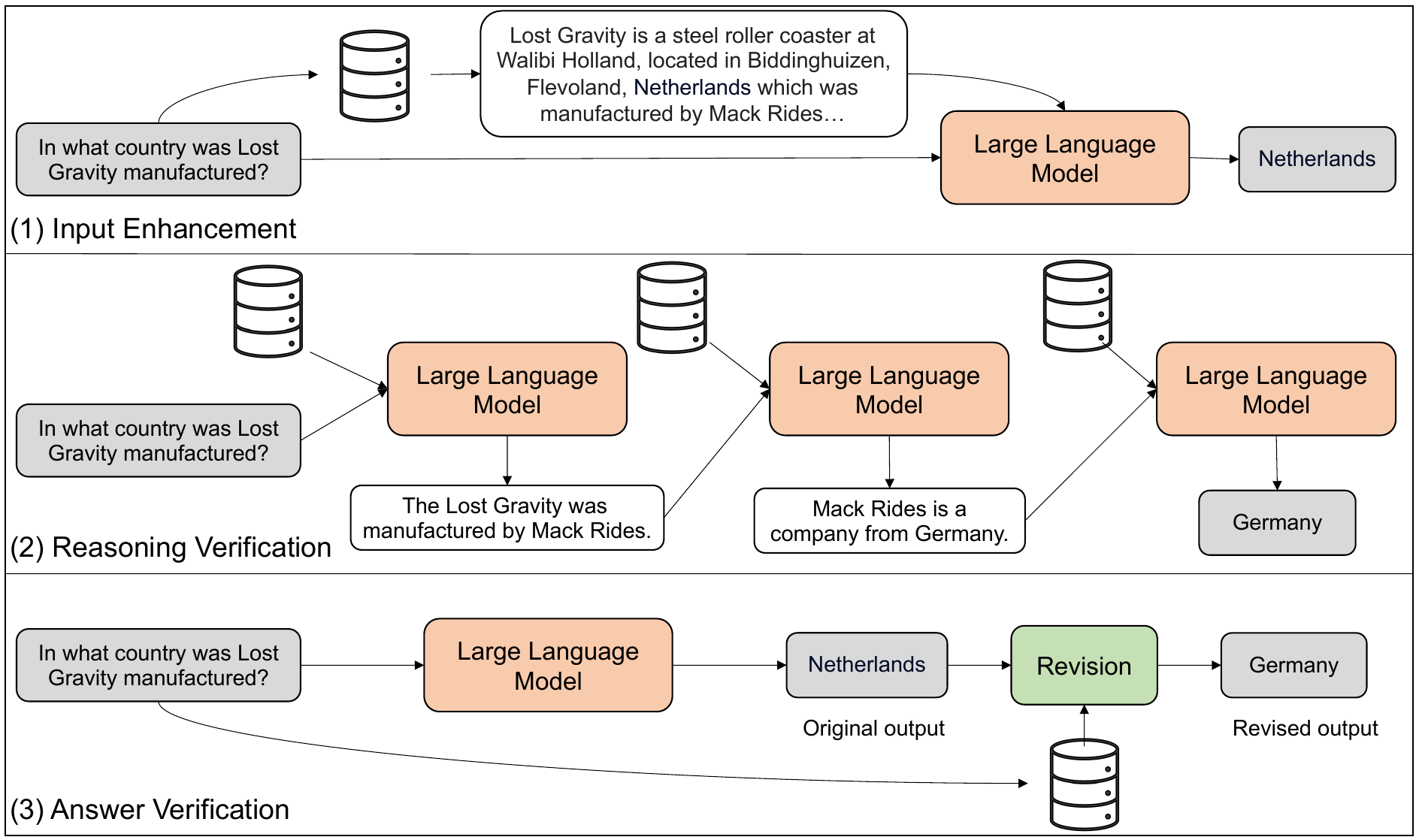}
  \caption{Three kinds of document utilization: (1) input enhancement uses relevant documents as part of the input prompt; (2) reasoning verification uses relevant documents to ensure that the reasoning process is correct; (3) answer verification modifies the original answers of large language models using relevant documents.}
  \label{fig:doc_util}
\end{figure*}

\paragraph{LLM-based.} 
Generative retrieval is a new paradigm of retrieval method that mainly includes two schemes: generating identifier strings of documents and generating completed documents.
The former uses identifiers to reduce the amount of useless information and make it easier for the model to memorize and learn \cite{li-etal-2023-multiview}.
\citet{de2020autoregressive} propose GENRE, which retrieves an entity by generating the entity text itself. GENRE also could be applied in page-level retrieval, where each document contains a unique title as the identifier.
\citet{lee2022generative} introduce generative retrieval to the multi-hop setting, and the retrieved items are short sentences.
\citet{tay2022transformer} propose the DSI method, which takes numeric IDs as identifiers for documents.
\citet{wang2022neural} improve the DSI by generating more queries as extra training data. However, the numeric Ids-based methods usually are evaluated on small datasets, partially because they suffer from the large scaling problem.
\citet{bevilacqua2022autoregressive} propose SEAL, which takes substrings as identifiers. The retrieval process is effectively completed upon the FM-Index structure.
\citet{li-etal-2023-multiview} propose multiview identifiers that represented a passage from different perspectives to enhance generative retrieval and achieve state-of-the-art performance.

Instead of generating identifiers, the latter aims to use large language models to directly generate complete documents.
Generate-then-read \cite{yu2023generate} shows that generated contextual documents contain the correct answer more often than the top retrieved documents and significantly outperform directly generating answers from large language models despite not incorporating any new external information.
RECITE~\cite{sun2023recitationaugmented} takes a similar approach, which tackles knowledge-intensive NLP tasks by first reciting relevant information and then generating the outputs.
PKG~\cite{luo2023augmented} equips LLMs with a background knowledge
generation module to access relevant knowledge.
The parametric knowledge module is based on open-source small language models and can be fine-tuned efficiently offline to store any knowledge.
\citet{feng2023cook} propose to empower general large language models with modular and collaboratively sourced knowledge through the integration of specialized language models.
The specialized language models are trained on corpora from diverse sources and domains.
They alo propose three levels of knowledge filters to dynamically select and refine generated documents and control for topic relevance, document brevity, and knowledge factuality.

\noindent
\textit{Highlight:} 
Retriever-based retrieval uses a retriever to fetch relevant documents from an external corpus.
However, the retrieved documents might contain noisy information that is irrelevant to the question.
Another option is to use large language models to directly generate relevant documents.
However, the latter method cannot obtain real-time information.
So we should make an appropriate choice according to the actual scenario.

\subsection{Document Utilization}
\label{sec:document_util}

Once we have the relevant documents, how can we use them to improve the capability of the large language models?
As shown in Figure~\ref{fig:doc_util}, we divide the different ways of using documents into three categories: input enhancement, reasoning verification, and answer verification.
\paragraph{Input Enhancement.}
Language language models define probability distributions over sequences of tokens.
The retrieved top-$k$ documents provide rich information about the original input context and can potentially help the LLMs to make a better prediction. 
A common approach is to prepend the retrieved documents to the input and feed them into large language models to make the final prediction~\cite{khattab2023demonstratesearchpredict, yu2023generate, luo2023augmented, feng2023knowledge}.

However, this simple scheme is fundamentally restricted by the number of documents we can include, given the language model's context window size.
REPLUG \cite{shi2023replug} introduces a new ensemble scheme that encodes the retrieved documents in parallel with the same black-box LM.
They prepend each document to input separately and then ensemble output probabilities from all $k$ passes.

\paragraph{Reasoning Verification.}
Large language models are capable of answering complex questions by generating step-by-step natural language reasoning steps — called chain of thought (CoT) \cite{wei2022chain}.
However, for many open-domain questions, all required knowledge is not always available or up-to-date in models’ parameters and it’s beneficial to retrieve knowledge from external sources \cite{lazaridou2022internet}.
IRCoT \cite{trivedi2023interleaving} proposes an interleaving approach to
use retrieval to guide the chain-of-thought reasoning steps and use CoT reasoning to guide the retrieval.
Self-ask \cite{press2023measuring} builds on chain of thought prompting instead of outputting a continuous undemarcated chain-of-thought.
Self-ask clearly demarcates the beginning and end of every sub-question and uses a search engine to answer the sub-questions instead of the LLMs. 
ReAct~\cite{yao2023react} prompts LLMs to generate both verbal reasoning traces and actions in an interleaved manner, which allows the model to perform dynamic reasoning to create, maintain, and adjust high-level plans for acting, while also interacting with the external environments to incorporate additional information into reasoning.
Verify-and-Edit~\cite{zhao2023verifyandedit} seeks to increase prediction factuality by post-editing reasoning chains according to external knowledge.
In addition,  several recent works~\cite{shao2023enhancing, feng2023retrieval, yu2023improving} first generate initial outputs, then utilize a retrieval model to acquire relevant information from large document collections, and finally incorporate the retrieved information into the in-context demonstration for output refinement.

\paragraph{Answer Verification.}
\citet{he2022rethinking} present a post-processing approach called rethinking with retrieval (RR) for utilizing external knowledge in LLMs.
They begin by using the chain-of-thought (CoT) prompting method \cite{wei2022chain} to generate a diverse set of reasoning paths.
They then use each reasoning step in those paths to retrieve relevant external knowledge, which enables RR to provide more faithful explanations and more accurate predictions.
Instead of constraining LMs to generate attributed text,  RARR \cite{gao-etal-2023-rarr} proposes a model-agnostic approach to improve the attribution of any existing LM.
After generating text, RARR fetches relevant evidence, and then revises the text to make it consistent with the evidence while preserving qualities like style or structure, enabling the revised text to be seamlessly used in place of the original. 
RARR can be viewed as a retrieval augmented model where retrieval happens after generation rather than before.
\citet{peng2023check} present LLM-AUGMENTER to improve LLMs with external knowledge and automated feedback.
Given a user query, LLM-AUGMENTER first retrieves evidence from external knowledge and further consolidates evidence by linking retrieved raw evidence with related context and performing reasoning to form evidence chains.
LLM-AUGMENTER then verifies the candidate's response by checking whether it hallucinates evidence.

\noindent
\textit{Highlight:} 
Input enhancement, reasoning verification, and answer verification are three common ways to use revelant documents.
They use the relevant documents in three different stages.
Input enhancement uses relevant documents as part of the input prompt.
Reasoning verification uses relevant documents to ensure that the reasoning process is correct.
Answer verification modifies the original answers of large language models using relevant documents.
It is worth exploring the effects of using related documents simultaneously at different stages.

\subsection{Knowledge Conflict}
\label{sec:knowledge_conflict}
\begin{figure}[!h]
  \includegraphics[width=\linewidth]{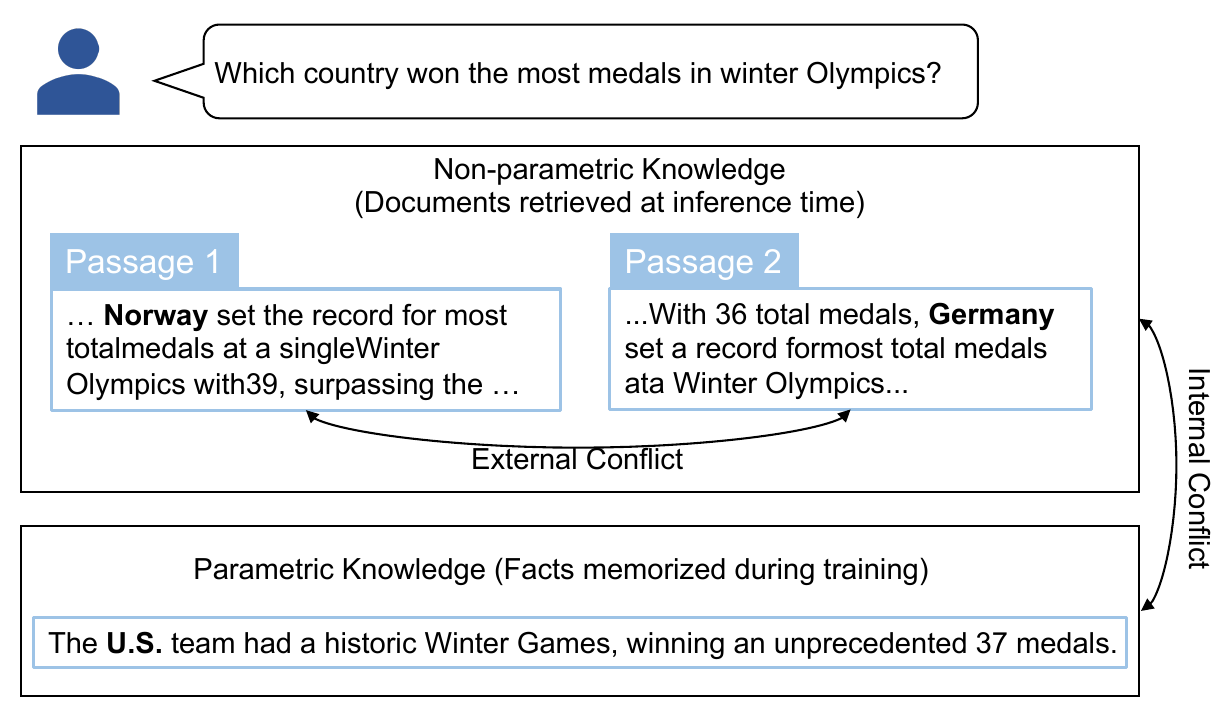}
  \caption{Two kinds of knowledge conflict: internal conflict:inconsistency between the knowledge in large language models and the knowledge in the retrieved documents, and external conflict: inconsistency between the retrieved multiple documents.}
  \label{fig:conflict}
\end{figure}
In retrieval-augmented LLMs, there are two sources of knowledge contributing to model inference with an ambiguous and opaque division of labor. 
The first is the implicit parametric knowledge (i.e., their learned weights) instilled by pre-training and fine-tuning.
The second is contextual knowledge, usually sourced as passages of text from the retriever.
Knowledge conflict means that the information contained is inconsistent and contradictory.
As shown in Figure~\ref{fig:conflict}, there are two types of knowledge conflicts: internal conflict and external conflict.
Internal conflict refers to the inconsistency between the knowledge in large language models and the knowledge in the retrieved documents.
External conflict refers to the inconsistency between the retrieved multiple documents.

\paragraph{Internal Conflict.}
As the world is constantly evolving, memorized facts may become outdated \cite{liska2022streamingqa, kasai2022realtime}.
Augment LLM prompting with external context containing relevant knowledge is a promising direction.
However, such methods face the challenge that LLMs may persist with the memorized facts and ignore the provided context \cite{longpre-etal-2021-entity}.
To tackle this challenge, recent
works \cite{neeman2022disentqa, li2022large} finetune LLMs on counterfactual contexts, where the original facts are replaced with counterfactual ones.
They find that such finetuning processes can effectively improve the LLMs’ utilization of contexts instead of relying solely on their parametric knowledge.
\citet{zhou2023contextfaithful} propose an approach using prompting to improve context faithfulness in LLMs without additional finetuning, which offers a
more general and cost-effective method for LLMs.
They present various prompting strategies to improve the faithfulness of LLMs, including designing effective prompts and choosing appropriate in-context demonstrations.
\citet{xie2023adaptive} present comprehensive and controlled investigation into the behavior of LLMs when encountering counter-memory.
They find that with both supportive and contradictory evidence to their parametric memory, LLMs show a strong confirmation bias and tend to cling to their parametric memory.

\paragraph{External Conflict.}
Such scenarios are common in settings where some passages are updated with new information, while other passages remain outdated \cite{zhang-choi-2021-situatedqa}.
Such conflicts can also occur when passages are adversarially edited to contain false information \cite{du2022synthetic}, or when passages are authored by multiple people who have differing opinions about an answer \cite{chen-etal-2019-seeing}.
\citet{chen2022rich} simulate a setting where a subset of evidence passages are perturbed to suggest a different answer to reflect the realistic scenario where retrieval returns a mixed bag of information. 
They find that when different passages suggest multiple conflicting answers, models prefer the answer that matches their parametric knowledge.
In addition to analyzing simple internal and external conflicts, \citet{xie2023adaptive} also experiment on more complicated knowledge conflict scenarios.
With both relevant and irrelevant evidence provided, LLMs can filter out the irrelevant ones to a certain extent. However, as the quantity of irrelevant evidence increases, such an ability diminishes.

\noindent
\textit{Highlight:} Knowledge conflict is a very important problem. 
However, current research focuses on the analysis of knowledge conflicts.
The next step should be to resolve knowledge conflicts from different aspects, such as data filtering, model fine-tuning.

\begin{table*}[t!]
    \centering
    \begin{tabular}{clc}
    \toprule
        \rowcolor[gray]{.92}
       \textbf{Task Type} & \textbf{DataSet} & \textbf{Metric} \\
        \midrule
            \multirow{3}{*}{Single-hop QA} 
                & Natural Questions \cite{kwiatkowski-etal-2019-natural} & EM / $\text{F}_1$ \\
                & TriviaQA \cite{joshi-etal-2017-triviaqa} &  EM / $\text{F}_1$ \\
                & PopQA~\cite{mallen-etal-2023-trust} &  Accuracy  \\ 
        \midrule
            \multirow{4}{*}{Multi-hop QA}
                & 2WikiMultiHopQA~\cite{ho-etal-2020-constructing}  & EM / $\text{F}_1$  \\
                & HotPotQA~\cite{yang-etal-2018-hotpotqa} &  EM / $\text{F}_1$ \\
                & MuSiQue~\cite{trivedi2022musique} & EM / $\text{F}_1$ \\
                & Bamboogle~\cite{press2023measuring} & EM / $\text{F}_1$ \\
        \midrule
            \multirow{3}{*}{Fact Verification}
                & Fever~\cite{thorne-etal-2018-fever} & Accuracy \\
                & Feverous~\cite{aly2021feverous} &  Accuracy \\
                & FoolMeTwice~\cite{eisenschlos-etal-2021-fool} & Accuracy \\  
        \midrule
            \multirow{6}{*}{Complex Reasoning}
                & StrategyQA~\cite{geva2021did} & Accuracy \\
                & CommonsenseQA~\cite{talmor-etal-2019-commonsenseqa} & Accuracy \\
                & CommonsenseQA2.0~\cite{talmor2022commonsenseqa} & Accuracy \\
                & CSQA~\cite{saha2018complex} & EM / ROUGE \\
                & TempQuestions~\cite{jia2018tempquestions}  & EM / $\text{F}_1$ \\
                & INFOTABS~\cite{gupta-etal-2020-infotabs} &  EM / $\text{F}_1$ \\  
        \bottomrule
    \end{tabular}
    \caption{Benchmarks for evaluating retrieval-augmented large language models.}
    \label{tab:retrieval_bechmark}
\end{table*}

\subsection{Benchmark}
\label{sec:benchmark}
Knowledge-sensitive tasks are ideal for evaluating retrieval-enhanced large language models because solving knowledge-sensitive tasks requires access to a large amount of information.
We investigate the commonly used datasets in detail and divide them into the following categories by task type.

\paragraph{Single-hop QA.} Single-hop questions have relatively simple structures and can be answered using information contained in the paragraph. 
There are several commonly used datasets, including Natural Questions (NQ), TriviaQA and PopQA.
Natural Questions~\cite{kwiatkowski-etal-2019-natural} consists of questions aggregated from the Google search engine, and the answers are annotated by human experts.
TriviaQA~\cite{joshi-etal-2017-triviaqa} consists of questions authored by trivia enthusiasts, and evidence documents are collected retrospectively from Wikipedia and the Web.
PopQA~\cite{mallen-etal-2023-trust} is a large-scale entity-centric QA dataset, which is constructed to sample more heavily from the tail and has significantly more low-popularity entities.

\paragraph{Multi-hop QA.} Multi-hop question answering is a challenging subfield of QA that involves answering questions that cannot be resolved with a direct answer from a single source or passage.
Models need to perform multiple steps of reasoning in order to answer a question.
There are several commonly used datasets, including
HotPotQA, 2WikiMultiHopQA, MuSiQue and Bamboogle.
HotPotQA~\cite{yang-etal-2018-hotpotqa} is collected by explicitly composing questions requiring reasoning about multiple supporting context documents.
2WikiMultihopQA~\cite{ho-etal-2020-constructing} is also constructed via composition, but they use a limited set of hand-authored compositional rules.
MuSiQue~\cite{trivedi2022musique} is constructed with a bottom-up process by carefully selecting and composing single-hop questions. 
With six composition structures, MuSiQue is more challenging and less cheatable than HotPotQA and 2WikiMultihopQA.
Bamboogle~\cite{press2023measuring} is a small dataset with 2-hop questions written by the authors, where all questions are sufficiently difficult to be unanswerable by a popular internet search engine, but where both supporting pieces of evidence can be found in Wikipedia.

\paragraph{Fact Verification.}
Fact verification, also called fact checking, is a challenging task that requires retrieving relevant evidence from plain text and use the evidence to verify given claims.
There are several commonly used datasets, including
Fever, Feverous and FoolMeTwice (FM2).
Fever~\cite{thorne-etal-2018-fever} is a large dataset for fact verification that requires retrieving sentence-level evidence to support if a claim is supported or refuted. 
In addition to unstructured text evidence, Feverous~\cite{aly2021feverous} also considers Wikipedia tables as a form of evidence.
The evidence retrieval in Feverous considers the entirety of a Wikipedia article and thus the evidence can be located in any section of the article except the reference sections.
Feverous is balanced, having almost an equal amount of instances containing, either exclusively text, tables, or both as evidence.
FoolMeTwice~\cite{eisenschlos-etal-2021-fool} is collected through a fun multi-player game, which encourages adversarial examples, drastically lowering the number of examples that can be solved using shortcuts compared to other datasets.

\paragraph{Complex Reasoning.}
Complex Reasoning includes different types of reasoning, such as commonsense reasoning, tabular reasoning, etc.
Commonsense reasoning is the foundation of human understanding, rooted in the basic knowledge and life experiences accumulated through daily life and social practice, which outlines practical knowledge of how the world works \cite{sap-etal-2020-commonsense}.
Commonsense reasoning tasks evaluate models’ reasoning skills in the physical world.
StrategyQA, CommonsenseQA and CommonsenseQA2.0 are widely used commonsense reasoning datasets.
StrategyQA~\cite{geva2021did} is a question-answering benchmark focusing on open domain questions where the required reasoning steps are implicit in the question and should be inferred using a strategy.
CommonsenseQA~\cite{talmor-etal-2019-commonsenseqa} and  CommonsenseQA2.0~\cite{talmor2022commonsenseqa}  is proposed to explore the commonsense understanding ability of large language models, which includes yes/no questions (or assertions) about everyday commonsense knowledge.
CSQA~\cite{saha2018complex} is a long-form QA, which aims to generate comprehensive answers to questions seeking complex information.
TempQuestions~\cite{jia2018tempquestions} is built to investigate temporal reasoning.
This dataset includes 1,271 temporal questions that are divided into four classes: explicit temporal, implicit temporal, temporal answer, and ordinal constraints.
INFOTABS~\cite{gupta-etal-2020-infotabs} consists of 23, 738 human-written textual hypotheses based on premises in the form of tables extracted from Wikipedia info-boxes.

\section{Frontier Applications}
\label{application}

\subsection{Knowledge editing}
Knowledge editing offers a cost-effective way of refreshing the outdated information in LLMs. Consequently, its primary purpose is to keep LLMs aligned with the continually evolving world. Towards the errors reported by users following deployment, sequential model editing (SME) also can stand out as an efficient method for rectifying a series of mistakes as a patching mechanism \cite{huang2023transformer,hartvigsen2022aging}. In addition to general applications, knowledge editing methods open up new avenues by focusing on information beyond factual knowledge. LLMs might output toxic text or leak personal information when subjected to adversarial prompts \cite{carlini2023aligned,qiu2023latent}. To mitigate this concern, \citet{patil2023sensitive} introduces an attack-and-defense framework based on knowledge editing methods to remove sensitive information from the model. In addition, compared to traditional controllable text generation techniques \cite{qian-etal-2022-controllable, gu-etal-2022-distributional}, knowledge editing can serve as a control methods for the generation of LLMs. From the social psychology, \citet{mao2023editing} innovatively employs knowledge editing methods to modify the personality for LLMs which can control the LLM's open view on specified topics and establish a benchmark dataset called PersonalityEdit. In summary, model editing will continue to play a significant role in the domains of model security and stance control.

\subsection{Retrieval Augmentation}
Augmenting language models with relevant information retrieved from various knowledge stores has been shown to be effective in improving performance on various knowledge-intensive tasks.
In open-domain question answering and fact verification, the model can answer the question more accurately by searching relevant documents in a large corpus or on the web.
In addition to classical natural language processing tasks, many new applications have emerged with the development of retrieval-augmented large language models.
LangChain \footnote{\url{https://www.langchain.com/}} is a 
powerful framework that provides a set of tools, components, and interfaces to simplify the process of creating applications powered by large language models and chat models. 
LangChain makes it easy to manage interactions with large language models, link multiple components together, and integrate additional resources.
ChatPDF \footnote{\url{https://www.chatpdf.com/}} is an AI tool that helps you understand and chat with PDF documents. It can identify key information, provide concise summaries, and answer your questions.
ChatDoctor~\cite{li2023chatdoctor} is an advanced language model that is specifically designed for medical applications.
Patients can interact with the ChatDoctor model through a chat interface, asking questions about their health, symptoms, or medical conditions. The model will then analyze the input and provide a response that is tailored to the patient's unique situation.
New Bing uses retrieval augmentation by combining ChatGPT with Microsoft’s search engine. 
New Bing Chat generates a search query from your prompt, retrieves relevant documents, and uses them as context for its results. 
New Bing also provides links to sources of information for the sentences it generates.
In summary, a large language model has more powerful knowledge understanding, and reasoning ability by retrieving relevant documents, and will have more application scenarios.
Baidu, iFlytek and Kunlun also offer similar services, such as ERNIE Bot~\footnote{\url{https://yiyan.baidu.com/}}, Spark~\footnote{\url{https://xinghuo.xfyun.cn/}} and Skywork~\footnote{\url{https://search.tiangong.cn}}.

\section{Future Directions}
\label{future work}

The development of knowledge editing and retrieval augmentation are still in a rudimentary stage and thus leaves much room for improvement. In this section, we offer a succinct overview of future research.

\paragraph{Multi-source Knowledge Augmentation.}
Existing knowledge enhancement methods mainly exhibit limitations in terms of the formats and varieties of incorporated knowledge. 
Most knowledge editing methods primarily center around factual knowledge represented as triples, thereby constraining the extent of modifiable knowledge \cite{DBLP:conf/nips/MengBAB22, DBLP:conf/iclr/MengSABB23}. 
Current retrieval augmentation methods mainly focus on unstructured text retrieval from Wikipedia or web \cite{shi2023replug, feng2023knowledge, vu2023freshllms}.
In real-world scenarios, a complex question may require fragmented evidence gathered from different sources to have the final answer.
It is worth exploring the impact of different sources and different formats of evidence for large language models.
In addition, it is extremely important to find a suitable way to integrate evidence from different sources.

\paragraph{Knowledge Augmented Multi-modal Large Language Models.}
Multi-modal learning has attracted increasing research attention due to is huge application potential, as a fundamental technique for vision-to-language reasoning~\cite{li2023blip,yu2023knowledge,chen2023measuring,yu2021learning}. 
How to endow the large language model with the ability of multi-modal reasoning is becoming a hot research topic. 
\citet{yu2021hybrid} propose to use extra multi-modal knowledge to augment language generation processing for reasoning. 
\citet{cheng2023edit} explore the application of current knowledge editing methods for refining multi-modal models and reveal that the effects remain further improved. 
RA-CM3~\cite{yasunaga2023retrievalaugmented} proposes a retrieval-augmented multi-modal model, which enables a base multi-modal model  to refer to relevant text and images fetched by a retriever from external memory. 
Future research can further investigate the integration of knowledge and multi-modal large language models to address complex challenges in the real world.

\paragraph{Large Language Model Based Agents.}
Autonomous agents have long been a prominent research focus in both academic and industry communities \cite{padgham2005developing}.
Through the acquisition of vast amounts of web knowledge, LLMs have demonstrated remarkable potential in achieving human-level intelligence and brought a glimmer of hope for the further
development of agents~\cite{openai2023gpt4, sumers2023cognitive}. 
These LLM-based agents can exhibit reasoning and planning abilities and  have been applied to various real-world scenarios~\cite{qian2023communicative, li2023camel}.
Due to the diversity of the real world, LLM-based agents need additional information to make decisions.
It is very important for the development of LLM-based Agents to explore the integration of knowledge and large language model methods in actual and complex scenarios.

\paragraph{Analysis of Knowledge Enhancement Methods.} 
Current knowledge enhancement methods mainly focus on the generation results of the model, and other aspects still need to be studied.
\citet{li2023unveiling} introduce innovative evaluation metrics to analyze the side effects of knowledge editing. 
In addition, \citet{pinter2023emptying} scrutinize the limitations of extensive models and the ramifications of knowledge editing, and raise concerns about the current methods, contending that they pose a potential risk to users of LLMs when compared to alternative interactive methods (e.g. retrieval-based architectures).
Future research can establish exhaustive and empirically-driven analysis, which will enhance the comprehension of the viability of current methods and guide the application of knowledge enhancement in actual scenarios.

\section{Conclusion}
\label{conclusion}

In this paper, we perform a survey on integration of knowledge and large language models and offer a broad view of its main directions, including knowledge editing and retrieval augmentation.
Moreover, we summarize the commonly used benchmarks and frontier applications and point out some promising research directions.
We hope this survey can offer readers a clear picture of the current progress and inspire more work.

\section*{Acknowledgements}
\label{Acknowledgements}
Xiaocheng Feng is the corresponding author of this work. We thank the anonymous reviewers for their insightful comments. This work was supported by the National Natural Science Foundation of China (NSFC) (grant 62276078, U22B2059), Nature Scientific Foundation of Heilongjiang Province(YQ2021F006), the Key R\&D Program of Heilongjiang via grant 2022ZX01A32, the International Cooperation Project of PCL, PCL2022D01 and the Fundamental Research Funds for the Central Universities (Grant No.HIT.OCEF.2023018).

\bibliography{anthology,custom}
\bibliographystyle{acl_natbib}




\end{document}

%% file: taxonomy.tex
\tikzstyle{my-box}=[
    rectangle,
    draw=hidden-draw,
    rounded corners,
    text opacity=1,
    minimum height=1.5em,
    minimum width=5em,
    inner sep=2pt,
    align=center,
    fill opacity=.5,
    line width=0.8pt,
]
\tikzstyle{leaf}=[my-box, minimum height=1.5em,
    text=black, align=left,font=\normalsize,
    inner xsep=2pt,
    inner ysep=4pt,
    line width=0.8pt,
]
\begin{figure*}[th!]
    \centering
    \resizebox{0.98\textwidth}{!}{
        \begin{forest}
            forked edges,
            for tree={
                grow=east,
                reversed=true,
                anchor=base west,
                parent anchor=east,
                child anchor=west,
                base=left,
                font=\large,
                rectangle,
                draw=hidden-draw,
                rounded corners,
                align=left,
                minimum width=4em,
                edge+={darkgray, line width=1pt},
                s sep=7pt,
                inner xsep=2pt,
                inner ysep=3pt,
                line width=0.8pt,
                ver/.style={rotate=90, child anchor=north, parent anchor=south, anchor=center},
            },
            where level=1{text width=6em,font=\normalsize,}{},
            where level=2{text width=6em,font=\normalsize,}{},
            where level=3{text width=9em,font=\normalsize,}{},
            where level=4{text width=5em,font=\normalsize,}{},
            [
                Integration of Knowledge and Large Language Models,fill=hidden-yellow!70,ver
                [
                    Knowledge \\ Editing (\S \ref{know_edit}),fill=hidden-blue!70
                    [
                     Input \\ Editing \\ (\S \ref{input_edit}),fill=hidden-blue!70
                           [
                                Prompt Augmented,fill=hidden-blue!70
                                [
                                    IKE~\cite{zheng2023can}\text{,} SuperICL~\cite{xu2023small}\text{,} \citet{yang2023chatgpt}\text{,} \\ PKG~\cite{luo2023augmented}\text{,} KAPING~\cite{baek2023knowledge}
                                    , leaf, text width=38em,fill=hidden-blue!70
                                ]
                           ]
                           [
                                Prompt Editing,fill=hidden-blue!70
                                [
                                    MemPrompt~\cite{madaan-etal-2022-memory}\text{,} PACE~\cite{dong2023pace}\text{,} MeLLo~\cite{zhong2023mquake}
                                    , leaf, text width=38em,fill=hidden-blue!70
                                ]
                           ]
                    ]
                    [
                     Model \\ Editing \\ (\S \ref{model_edit}),fill=hidden-blue!70
                            [
                                Knowledge Plug-in,fill=hidden-blue!70
                                [
                                    NKB~\cite{dai2022neural}\text{,} CALINET~\cite{dong-etal-2022-calibrating}\text{,}                           SERAC~\cite{mitchell2022memory}\text{,}\\T-Patcher~\cite{huang2023transformer}\text{,} GRACE~\cite{hartvigsen2022aging}
                                    , leaf, text width=38em,fill=hidden-blue!70
                                ]
                            ]
                            [
                                Locate-then-Edit,fill=hidden-blue!70
                                [
                                     KN~\cite{dai2022knowledge}\text{,} ROME~\cite{DBLP:conf/nips/MengBAB22}\text{,} MEMIT~\cite{DBLP:conf/iclr/MengSABB23}\text{,} \\
                                     PMET~\cite{DBLP:journals/corr/abs-2308-08742}\text{,} \citet{DBLP:journals/corr/abs-2305-14956}\text{,}
                                    , leaf, text width=38em,fill=hidden-blue!70
                                ]
                            ]
                            [
                                Overall Editing,fill=hidden-blue!70
                                [
                                 KE~\cite{de-cao-etal-2021-editing}\text{,} MEND~\cite{DBLP:conf/iclr/MitchellLBFM22}\text{,} \citet{DBLP:conf/iclr/IlharcoRWSHF23}
                                    , leaf, text width=38em,fill=hidden-blue!70
                                ]
                            ]
                    ]
                    [
                     Assess \\ Knowledge \\ Editing \\ (\S \ref{knowledge_edit_assess}),fill=hidden-blue!70
                            [
                                MQUAKE~\cite{zhong2023mquake}\text{,} RippleEdits~\cite{cohen2023evaluating}\text{,} \citet{brown2023edit}\text{,}\\ Eva-KELLM~\cite{DBLP:journals/corr/abs-2308-09954}\text{,} CounterFact+~\cite{hoelscherobermaier2023detecting}\text{,}                         KLoB~\cite{ju2023klob}
                                , leaf, text width=48.5em,fill=hidden-blue!70
                            ]
                    ]
                ]
                [
                    Retrieval \\ Augmentation \\ (\S \ref{sec:retrieval_aug}),fill=hidden-red!70
                    [
                        Retrieval \\ Judgement \\  (\S \ref{sec:retrieval_judge}),fill=hidden-red!70
                        [
                            Calibration-based ,fill=hidden-red!70
                            [
                                FLARE~\cite{jiang2023active}{, }\citet{mallen-etal-2023-trust}{,}\citet{kandpal2023large}{,}\citet{jiang-etal-2021-know}
                                , leaf, text width=38em,fill=hidden-red!70
                            ]
                        ]
                        [
                            Model-based ,fill=hidden-red!70
                            [
                                \citet{ren2023investigating}{, }\citet{Yoran_Wolfson_Ram_Berant_2023}{, }\citet{kadavath2022language}{, }\citet{yin-etal-2023-large}{, } \\ \citet{feng2023knowledge}{, } \citet{baek2023knowledgeaugmented}
                                    , leaf, text width=38em,fill=hidden-red!70
                            ]
                        ]
                    ]
                    [
                        Document \\ Retrieval  \\ (\S \ref{sec:retrieval_method}),fill=hidden-red!70
                        [
                            Retriever-based,fill=hidden-red!70
                            [
                                BM25~\cite{robertson2009probabilistic}{, } DPR~\cite{DBLP:conf/emnlp/KarpukhinOMLWEC20}{, } RAG~\cite{lewis2021retrievalaugmented}{, } \\RETRO~\cite{borgeaud2022improving}{, } Atlas~\cite{izacard2022atlas}{, } AAR~\cite{yu-etal-2023-augmentation}{, } \\ REPLUG LSR~\cite{shi2023replug}{, } Query2doc~\cite{wang2023query2doc}{, } \citet{ma2023query}
                                , leaf, text width=38em,fill=hidden-red!70
                            ]
                        ]
                        [
                            LLM-based,fill=hidden-red!70
                            [
                              GENRE~\cite{de2020autoregressive}{, } \citet{lee2022generative}{, } DSI~\cite{tay2022transformer}{, } \\ SEAL~\cite{bevilacqua2022autoregressive}{, } Multiview~\cite{li-etal-2023-multiview}{, } PKG~\cite{luo2023augmented}{, } \\ GENREAD \cite{yu2023generate}{, } RECITE~\cite{sun2023recitationaugmented}{, } COOK~\cite{feng2023cook}
                                , leaf, text width=38em,fill=hidden-red!70
                            ]
                        ]    
                    ]
                    [
                        Document \\ Utilization \\ (\S \ref{sec:document_util}),fill=hidden-red!70
                       [
                            Input Enhancement,fill=hidden-red!70
                            [
                                REPLUG~\cite{shi2023replug}{, } RECITE~\cite{sun2023recitationaugmented}{, } PKG~\cite{luo2023augmented}{, } \\ GENREAD~\cite{yu2023generate}{, } \citet{ram2023incontext}{, }
                                 , leaf, text width=38em,fill=hidden-red!70
                            ]
                       ]
                       [
                            Reasoning  Verification,fill=hidden-red!70
                            [
                                Self-ask~\cite{press2023measuring}{, } IRCoT~\cite{trivedi2023interleaving}{,}  ReAct~\cite{yao2023react}{, } \\ ITER-RETGEN~\cite{shao2023enhancing}{,} ITRG~\cite{feng2023retrieval}{, }  \\ REFEED~\cite{yu2023improving}{, } Verify-and-Edit~\cite{zhao2023verifyandedit}
                                 , leaf, text width=38em,fill=hidden-red!70
                            ]
                       ]
                       [
                            Answer Verification,fill=hidden-red!70
                            [
                                RARR~\cite{gao-etal-2023-rarr}{, } LLM-AUGMENTER~\cite{peng2023check}{, } \citet{he2022rethinking}
                                 , leaf, text width=38em,fill=hidden-red!70
                            ]
                       ]
                    ]
                    [
                        Knowledge \\ Conflict  \\ 
                        (\S \ref{sec:knowledge_conflict}),fill=hidden-red!70
                        [
                            Internal Conflict,fill=hidden-red!70
                            [
                                 \citet{longpre-etal-2021-entity}{, } \citet{neeman2022disentqa}{, } \citet{zhou2023contextfaithful}{, }
                                 \citet{xie2023adaptive}
                                , leaf, text width=38em,fill=hidden-red!70
                            ]
                        ]
                        [
                            External Conflict,fill=hidden-red!70
                            [
                             \citet{du2022synthetic}{, } \citet{chen-etal-2019-seeing}{, } \citet{chen2022rich}{, } \citet{xie2023adaptive}
                            , leaf, text width=38em,fill=hidden-red!70
                            ]
                        ]
                    ]
                    [
                        Benchmark \\ (\S \ref{sec:benchmark}),fill=hidden-red!70
                        [
                            Single-hop QA,fill=hidden-red!70
                            [
                                Natural Questions~\cite{kwiatkowski-etal-2019-natural}{, } TriviaQA~\cite{joshi-etal-2017-triviaqa}{, } \\ PopQA~\cite{mallen-etal-2023-trust}
                                 , leaf, text width=38em,fill=hidden-red!70
                            ]
                        ]
                        [
                            Multi-hop QA,fill=hidden-red!70
                            [
                               2WikiMultiHopQA~\cite{ho-etal-2020-constructing}{, } HotPotQA~\cite{yang-etal-2018-hotpotqa}{, } \\ MuSiQue~\cite{trivedi2022musique}{, } Bamboogle~\cite{press2023measuring}
                                 , leaf, text width=38em ,fill=hidden-red!70
                            ]
                        ]
                        [
                            Fact Verification,fill=hidden-red!70
                            [
                              Fever~\cite{thorne-etal-2018-fever}{, } Feverous~\cite{aly2021feverous}{, } \\ FoolMeTwice~\cite{eisenschlos-etal-2021-fool}
                                 , leaf, text width=38em ,fill=hidden-red!70
                            ]
                        ]
                        [
                            Complex Reasoning,fill=hidden-red!70
                            [
                                StrategyQA~\cite{geva2021did}{, }
                                CommonsenseQA~\cite{talmor-etal-2019-commonsenseqa}{, } \\ CommonsenseQA2.0~\cite{talmor2022commonsenseqa}{, } CSQA~\cite{saha2018complex}{, } \\ TempQuestions~\cite{jia2018tempquestions}{, }
                                INFOTABS~\cite{gupta-etal-2020-infotabs}
                                 , leaf, text width=38em ,fill=hidden-red!70
                            ]
                        ]
                    ]
                ]
            ]
        \end{forest}
    }
    \caption{Taxonomy of trends in integration of knowledge and large language models.}
    \label{fig:survey}
\end{figure*}

%% file: knowledge_edit.tex
\subsection{Input Editing}
\label{input_edit}
The extensive parameter scale and "black-box" form of numerous large models often impede them from undergoing fine-tuning commonly for the acquisition of new knowledge, such as ChatGPT, Bard\footnote{\url{https://bard.google.com/}}. Therefore, the most direct approach to infusing knowledge into LLMs involves editing the input \cite{zheng2023can,luo2023augmented}, which incurs minimal costs and resource requirements. There are two aspects of input editing: the inclusion of external information to enhance prompts, and editing prompts based on feedback. Adjusting input not only offers an intuitive and comprehensible depiction of the process of new knowledge but also guarantees the preservation of the original model's knowledge.

\paragraph{Prompt Augmented.} In-context learning (ICL) has been proven a useful paradigm for LLMs based on a few demonstrations that are tailored for diverse tasks. Based on the goal of knowledge editing, IKE \cite{zheng2023can} designs three different types of prompts: copy, update, and retain to enhance the generalization of newly injected knowledge and keep original information in LLMs. SuperICL \cite{xu2023small} introduces a smaller model fine-tuned on task-specific data as a plug-in for LLMs to augment the conventional ICL method. The smaller model produces the predicted label and confidence score for each original demonstration which assists the LLM in grasping the complexity of the given examples. During the inference phase, the LLM generates a final prediction for the test input which comprises the reconstructed context, input text, and plug-in model's prediction. \citet{luo2023augmented} similarly train a smaller model as an auxiliary parametric knowledge guiding (PKG) framework to generate background documents for domain-specific tasks enhancing the prompt. In addition, leveraging knowledge graphs (KGs) to augment pre-trained language models has become a prevalent practice \cite{yang2023chatgpt,moiseev-etal-2022-skill}, some methods expand the input using knowledge graphs as an external resource without training. For example, KAPING \cite{baek2023knowledge} retrieves the relevant facts from KG and integrates them with the original question as a new prompt to generate the answer. \citet{andrus2022enhanced} propose an architecture aiming at enhancing story comprehension for LLMs. Initially, it extracts entities from the story document to construct KGs. Subsequently, question-related triples are also retrieved and transformed into natural language sentences to create an informative prompt. 

\noindent 
\textit{Highlight:} These methods mainly concentrate on incorporating new facts into prompts using diverse approaches, which are relatively straightforward and highly achievable. However, they exhibit a limited capacity to correct errors in LLMs.

\paragraph{Prompt Editing.} In addition to introducing background information into the prompt, the process of deconstructing and refining the prompt can also contribute to enhancing the accuracy of the LLMs' responses. MemPrompt \cite{madaan-etal-2022-memory} is designed by incorporating user feedback to mitigate instances where the LLM misunderstood prompts. It establishes an expanding memory to retain users' past assessments of the model's understanding of the task. Subsequently, the architecture produces an improved prompt by combining it with user feedback retrieved from this memory resource. In contrast to the utilization of user feedback, PACE \cite{dong2023pace} leverages LLMs' own judgment to criticize the response and refine the prompt, which successfully boosts the performance of medium/low-quality human-written prompts. To enhance the generality of injected knowledge for multi-hop questions, MeLLo \cite{zhong2023mquake} decomposes the original question into sub-questions. This approach based on a self-checking mechanism, empowers the LLM to adapt the output of the sub-questions according to the retrieved facts from an explicit memory. 

\noindent 
\textit{Highlight:} It is noteworthy that editing prompts can facilitate LLMs in accurately tackling complicated reasoning tasks and rectifying errors, making it particularly suitable in "black-box" scenarios. Nevertheless, these methods introduce facts in a disposable manner, without fundamentally modifying the intrinsic knowledge within LLMs.

\subsection{Model Editing}
\label{model_edit}
Instead of editing input, numerous works are dedicated to fine-grained model editing in a parametric fashion, which can ensure the persistence of injected knowledge. According to different operations targeting LLMs' parameters, we categorize them into three classes, namely knowledge plug-in, locate-then-edit, and overall editing.

\paragraph{Knowledge Plug-in.} This paradigm introduces external plug-in parametric knowledge to edit LLMs, without altering the original weights. Following the perspective that knowledge resides within MLP layers \cite{geva-etal-2021-transformer}, NKB \cite{dai2022neural} and CALINET \cite{dong-etal-2022-calibrating} both adjust the output of the feed-forward-network (FFN) by adding additional memory slots. SERAC \cite{mitchell2022memory} maintains a scope classifier that determines whether the inputs fall within the scope of each edit stored in an external edit memory. Inputs deemed to be in scope will be passed to a counterfactual model along with the associated edit examples, while out-of-scope inputs will be processed by the original model. In addition to the traditional one-step editing scenario, plug-in knowledge is also applied to sequential model editing tasks suitable for practical situations. T-Patcher \cite{huang2023transformer} incorporates an extra trainable patch into the last FFN layer for each mistake. This mechanism allows the model to train these patches to correct errors while keeping the original parameters frozen. GRACE \cite{hartvigsen2022aging} implements sequential editing by adding an adapter at a selected layer. This adapter contains a discrete codebook designed to map activations to the corresponding values, along with a deferral mechanism that achieves codebook updates to encourage similar edits with the same values. The frame enables efficient model editing with low impact on unrelated inputs.

\noindent 
\textit{Highlight:} These methods introduce plug-in knowledge to quickly complete model editing in lower resource scenes, but they often cannot achieve precise editing and the added objects are mostly in the last FFN layer of the model.

\paragraph{Locate-then-Edit.} This paradigm employs a dual-stage framework to edit the model in a fine-grained manner. It first locates the specific parameters about new facts and makes targeted modifications to inject this knowledge. \citet{dai2022knowledge} adapt a method on integrated gradients to evaluate the contribution of each neuron in the second linear layer of FFN to knowledge predictions and then selects the knowledge neurons (KN) to update. ROME \cite{DBLP:conf/nips/MengBAB22} focuses on GPT-like autoregressive models, which introduces corruption to the embeddings of the subject token and successively restore internal activations to their clean value to acquire the causal importance for MLP layers. This approach will compute the new key-value pair vector which is inserted into the original matrix in specific FFN for each fact to edit the model precisely. However, most model editing methods can only inject a single piece of knowledge at a time. MEMIT \cite{DBLP:conf/iclr/MengSABB23} based on ROME defines an new expansion objective solved by the normal equations for updating thousands of edits at once. Instead of modifying a single layer, MEMIT spreads updates from a set of key-value pairs over the identified layers to improve the robustness. As finding multi-head self-attention (MHSA) works as a knowledge extractor in transformer component, PMET \cite{DBLP:journals/corr/abs-2308-08742} concurrently optimizes the hidden states of MHSA and FFN, surpassing the performance of MEMIT. Moreover, in accordance with the specific attributes of distinct knowledge domains, the model editing approach is tailored accordingly. \citet{DBLP:journals/corr/abs-2305-14956} adapt the MEMIT method to edit commonsense knowledge by broadening the scope of corruption tokens to encompass subjects, objects, and verbs.

\noindent 
\textit{Highlight:} These methods require an additional step to identify the parameters that store knowledge about edits. However, \citet{DBLP:journals/corr/abs-2301-04213} demonstrates that these knowledge-storage parameters are not exactly the same as the neurons that can be efficiently adjusted to edit the model. Therefore, the selection of appropriate parameters plays a significant role in model editing.

\paragraph{Overall Editing.} This paradigm involves editing the model by directly modifying it, bypassing the need to locate the specific parameters where knowledge is stored. A common approach is to train a hyper-network based on meta-learning to learn the model weights on new facts. Knowledge editor (KE) \cite{de-cao-etal-2021-editing} trains a bidirectional-LSTM as a hyper-network with constrained optimizations, to predict the updates required for injecting each atomic fact. However, this method cannot be applied to large models and computationally infeasible when editing knowledge. MEND \cite{DBLP:conf/iclr/MitchellLBFM22} introduces a low-rank decomposition of the gradients to optimize the process of hyper-network learning the parameter update from standard fine-tuning. This approach is capable of training small auxiliary networks with limited resources which effectively edits the behavior of LLMs. In addition to meta-learning, \citet{DBLP:conf/iclr/IlharcoRWSHF23} propose a method to simply edit models through task arithmetic. The framework illustrates that by combining task vectors derived from the element-wise difference between parameters after pre-training and fine-tuning, becomes possible to control the performance of models on various tasks. 

\noindent 
\textit{Highlight:} These methods are careless about the internals of the model, and obtain the update of model editing in a data-driven manner, without being constrained to a particular model. Nevertheless, in the case of LLMs, these approaches often require many additional cost, particularly in the context of hyper-network methods, which demand their own parameter size reduction in careful design.

\input{tables/knowledge_editing_benchmark}

\subsection{Assess Knowledge Editing}
\label{knowledge_edit_assess}
After editing the input and model, assessing the extent of knowledge integration can be accomplished by scrutinizing the output. This subsection will primarily introduce the characteristics of model evaluation and provide an overview of general benchmarks for knowledge editing in Table \ref{tab:eval_bechmark}.
 
The current methods for editing knowledge mainly aim at incorporating triple-fact knowledge, which concentrate on question-answer (QA) tasks,i.e., ZsRE \cite{levy-etal-2017-zero}. In addition, CounterFact is an evaluation dataset specially constructed for knowledge editing tasks to measure the effectiveness of significant changes in comparison to merely superficial alterations of target words \cite{DBLP:conf/nips/MengBAB22}. There are three main properties of assessing knowledge editing accounting for \textit{reliability}, \textit{generality}, \textit{locality} \cite{yao2023editing,huang2023transformer}.

\paragraph{Reliability.} The primary objective of the post-edited model is to generate the desired predictions for edited input \cite{de-cao-etal-2021-editing}. The reliability which is directly evaluated using the datasets mentioned, is assessed based on the success rate in knowledge editing \cite{levy-etal-2017-zero,DBLP:conf/nips/MengBAB22}.

\paragraph{Generality.} The generality is reflected in the fact that the post-edited model can successfully update the relevant facts of the edits, which includes numerous aspects. In addition to fundamental semantic transcription like sentence rephrasing and back translation, there are other more comprehensive assessment methods. \citet{yao2023editing} introduce a new metric called \textit{portability} which assesses the robustness of generalization in order to ascertain whether the post-edit model merely memorizes the superficial alterations in wording. MQUAKE serves as a challenging benchmark to evaluate whether the edited model can correctly and synchronously update the answers to multi-hop questions related to the edits \cite{zhong2023mquake}. RippleEdits also focuses on the ripple effects of knowledge editing based on six evaluation criteria, most of which assess the generality of injected knowledge within a 2-hop distance from edited facts \cite{cohen2023evaluating}.

\paragraph{Locality.} Determining whether the edited model retains unrelated knowledge of the edited fact is also a crucial property of the evaluation process. There are also two criteria to evaluate the locality of ripple effects of knowledge editing in RippleEdits \cite{cohen2023evaluating}. CounterFact+ is a more sensitive benchmark extending the original Counterfact to detect the unwanted side effects brought by knowledge editing, which influences unrelated facts and the next token probability distribution \cite{hoelscherobermaier2023detecting}.

Besides the aforementioned evaluation methods, several studies have raised attention towards examining a special kind of generalization: the cross-lingual capabilities of knowledge editing methods. Bi-ZsRE \cite{wang2023crosslingual} constructs samples in both the Chinese and English languages, while also assessing traditional editing methods. The findings indicate that current methods struggle to effectively transfer edited knowledge from one language to another. In addition, Eva-KELLM \cite{DBLP:journals/corr/abs-2308-09954} also evaluates the cross-language capabilities of knowledge editing, which pioneers the direct editing of raw documents for greater convenience. 

~\\
\noindent 
\textit{Highlight:} Aside from assessing the based effects, there are profound analyses to comprehensively evaluate knowledge editing methods. KLoB \cite{ju2023klob} meticulously emphasizes knowledge-locating methods based on three crucial assessing criteria: consistency, relevance, and unbiasedness. \citet{brown2023edit} explore the robustness of the post-edited model, which experimentally identifies the degrades of general robustness from knowledge editing. These thorough evaluations facilitate the selection of appropriate editing methods in real-world deployment scenarios.

%% file: tables/knowledge_editing_benchmark.tex
\begin{table*}[t!]
    \small
    \centering
    \begin{tabular}{ccccc}
    \toprule
        \rowcolor[gray]{.92} \textbf{DataSet} & \textbf{Language} & \textbf{Size} & \textbf{Evaluation} \\ 
        \midrule
            \makecell{ZsRE\\ \cite{levy-etal-2017-zero}} & En & 182,282 & Reliability/Generality/Locality \\
            \makecell{Counterful\\ \cite{DBLP:conf/nips/MengBAB22}} & En & 21,919 & Reliability/Generality/Locality \\
            \makecell{Counterful+\\ \cite{hoelscherobermaier2023detecting}} & En & 21,919 & Reliability/Generality/Locality  \\
            \makecell{Bi-ZsRE\\ \cite{wang2023crosslingual}} & En\&Zh & 14037\&14037 & Reliability/Generality/Locality/Portability/Cross-Lingual   \\
            \makecell{MQUAKE\\ \cite{zhong2023mquake}} & En & 11,043 & Generality  \\
            \makecell{RippleEdits\\ \cite{cohen2023evaluating}} & En & 4,000 & Reliability/Generality/Locality  \\
            \makecell{Eva-KELLM\\ \cite{DBLP:journals/corr/abs-2308-09954}} & En\&Zh & 8,882\&6,930 & Reliability/Generality/Locality/Cross-Lingual  \\
        \bottomrule
    \end{tabular}
    \caption{Comparison of knowledge editing evaluation benchmarks, including language, size and evaluation.  
    }
    \label{tab:eval_bechmark}
\end{table*}

%% file: acl_latex.bbl
\begin{thebibliography}{147}
\expandafter\ifx\csname natexlab\endcsname\relax\def\natexlab#1{#1}\fi

\bibitem[{Aly et~al.(2021)Aly, Guo, Schlichtkrull, Thorne, Vlachos,
  Christodoulopoulos, Cocarascu, and Mittal}]{aly2021feverous}
Rami Aly, Zhijiang Guo, Michael Schlichtkrull, James Thorne, Andreas Vlachos,
  Christos Christodoulopoulos, Oana Cocarascu, and Arpit Mittal. 2021.
\newblock Feverous: Fact extraction and verification over unstructured and
  structured information.
\newblock \emph{arXiv preprint arXiv:2106.05707}.

\bibitem[{Andrus et~al.(2022)Andrus, Nasiri, Cui, Cullen, and
  Fulda}]{andrus2022enhanced}
Berkeley~R Andrus, Yeganeh Nasiri, Shilong Cui, Benjamin Cullen, and Nancy
  Fulda. 2022.
\newblock Enhanced story comprehension for large language models through
  dynamic document-based knowledge graphs.
\newblock In \emph{Proceedings of the AAAI Conference on Artificial
  Intelligence}, volume~36, pages 10436--10444.

\bibitem[{Baek et~al.(2023{\natexlab{a}})Baek, Aji, and
  Saffari}]{baek2023knowledge}
Jinheon Baek, Alham~Fikri Aji, and Amir Saffari. 2023{\natexlab{a}}.
\newblock Knowledge-augmented language model prompting for zero-shot knowledge
  graph question answering.
\newblock \emph{arXiv preprint arXiv:2306.04136}.

\bibitem[{Baek et~al.(2023{\natexlab{b}})Baek, Jeong, Kang, Park, and
  Hwang}]{baek2023knowledgeaugmented}
Jinheon Baek, Soyeong Jeong, Minki Kang, Jong~C. Park, and Sung~Ju Hwang.
  2023{\natexlab{b}}.
\newblock \href {http://arxiv.org/abs/2310.12836} {Knowledge-augmented language
  model verification}.

\bibitem[{Bevilacqua et~al.(2022)Bevilacqua, Ottaviano, Lewis, Yih, Riedel, and
  Petroni}]{bevilacqua2022autoregressive}
Michele Bevilacqua, Giuseppe Ottaviano, Patrick Lewis, Scott Yih, Sebastian
  Riedel, and Fabio Petroni. 2022.
\newblock Autoregressive search engines: Generating substrings as document
  identifiers.
\newblock \emph{Advances in Neural Information Processing Systems},
  35:31668--31683.

\bibitem[{Borgeaud et~al.(2022)Borgeaud, Mensch, Hoffmann, Cai, Rutherford,
  Millican, van~den Driessche, Lespiau, Damoc, Clark, de~Las~Casas, Guy,
  Menick, Ring, Hennigan, Huang, Maggiore, Jones, Cassirer, Brock, Paganini,
  Irving, Vinyals, Osindero, Simonyan, Rae, Elsen, and
  Sifre}]{borgeaud2022improving}
Sebastian Borgeaud, Arthur Mensch, Jordan Hoffmann, Trevor Cai, Eliza
  Rutherford, Katie Millican, George van~den Driessche, Jean-Baptiste Lespiau,
  Bogdan Damoc, Aidan Clark, Diego de~Las~Casas, Aurelia Guy, Jacob Menick,
  Roman Ring, Tom Hennigan, Saffron Huang, Loren Maggiore, Chris Jones, Albin
  Cassirer, Andy Brock, Michela Paganini, Geoffrey Irving, Oriol Vinyals, Simon
  Osindero, Karen Simonyan, Jack~W. Rae, Erich Elsen, and Laurent Sifre. 2022.
\newblock \href {http://arxiv.org/abs/2112.04426} {Improving language models by
  retrieving from trillions of tokens}.

\bibitem[{Bowman et~al.(2015)Bowman, Angeli, Potts, and
  Manning}]{bowman2015large}
Samuel~R Bowman, Gabor Angeli, Christopher Potts, and Christopher~D Manning.
  2015.
\newblock A large annotated corpus for learning natural language inference.
\newblock \emph{arXiv preprint arXiv:1508.05326}.

\bibitem[{Brown et~al.(2023)Brown, Godfrey, Nizinski, Tu, and
  Kvinge}]{brown2023edit}
Davis Brown, Charles Godfrey, Cody Nizinski, Jonathan Tu, and Henry Kvinge.
  2023.
\newblock \href {http://arxiv.org/abs/2303.00046} {Edit at your own risk:
  evaluating the robustness of edited models to distribution shifts}.

\bibitem[{Brown et~al.(2020)Brown, Mann, Ryder, Subbiah, Kaplan, Dhariwal,
  Neelakantan, Shyam, Sastry, Askell, Agarwal, Herbert-Voss, Krueger, Henighan,
  Child, Ramesh, Ziegler, Wu, Winter, Hesse, Chen, Sigler, Litwin, Gray, Chess,
  Clark, Berner, McCandlish, Radford, Sutskever, and
  Amodei}]{brown2020language}
Tom~B. Brown, Benjamin Mann, Nick Ryder, Melanie Subbiah, Jared Kaplan,
  Prafulla Dhariwal, Arvind Neelakantan, Pranav Shyam, Girish Sastry, Amanda
  Askell, Sandhini Agarwal, Ariel Herbert-Voss, Gretchen Krueger, Tom Henighan,
  Rewon Child, Aditya Ramesh, Daniel~M. Ziegler, Jeffrey Wu, Clemens Winter,
  Christopher Hesse, Mark Chen, Eric Sigler, Mateusz Litwin, Scott Gray,
  Benjamin Chess, Jack Clark, Christopher Berner, Sam McCandlish, Alec Radford,
  Ilya Sutskever, and Dario Amodei. 2020.
\newblock \href {http://arxiv.org/abs/2005.14165} {Language models are few-shot
  learners}.

\bibitem[{Carlini et~al.(2023)Carlini, Nasr, Choquette-Choo, Jagielski, Gao,
  Awadalla, Koh, Ippolito, Lee, Tramer, and Schmidt}]{carlini2023aligned}
Nicholas Carlini, Milad Nasr, Christopher~A. Choquette-Choo, Matthew Jagielski,
  Irena Gao, Anas Awadalla, Pang~Wei Koh, Daphne Ippolito, Katherine Lee,
  Florian Tramer, and Ludwig Schmidt. 2023.
\newblock \href {http://arxiv.org/abs/2306.15447} {Are aligned neural networks
  adversarially aligned?}

\bibitem[{Chen et~al.(2022)Chen, Zhang, and Choi}]{chen2022rich}
Hung-Ting Chen, Michael~JQ Zhang, and Eunsol Choi. 2022.
\newblock Rich knowledge sources bring complex knowledge conflicts:
  Recalibrating models to reflect conflicting evidence.
\newblock \emph{arXiv preprint arXiv:2210.13701}.

\bibitem[{Chen et~al.(2019)Chen, Khashabi, Yin, Callison-Burch, and
  Roth}]{chen-etal-2019-seeing}
Sihao Chen, Daniel Khashabi, Wenpeng Yin, Chris Callison-Burch, and Dan Roth.
  2019.
\newblock \href {https://doi.org/10.18653/v1/N19-1053} {Seeing things from a
  different angle:discovering diverse perspectives about claims}.
\newblock In \emph{Proceedings of the 2019 Conference of the North {A}merican
  Chapter of the Association for Computational Linguistics: Human Language
  Technologies, Volume 1 (Long and Short Papers)}, pages 542--557, Minneapolis,
  Minnesota. Association for Computational Linguistics.

\bibitem[{Chen et~al.(2023)Chen, Sikka, Cogswell, Ji, and
  Divakaran}]{chen2023measuring}
Yangyi Chen, Karan Sikka, Michael Cogswell, Heng Ji, and Ajay Divakaran. 2023.
\newblock Measuring and improving chain-of-thought reasoning in vision-language
  models.
\newblock \emph{arXiv preprint arXiv:2309.04461}.

\bibitem[{Cheng et~al.(2023)Cheng, Tian, Liu, Chen, Wang, Chen, and
  Zhang}]{cheng2023edit}
Siyuan Cheng, Bozhong Tian, Qingbin Liu, Xi~Chen, Yongheng Wang, Huajun Chen,
  and Ningyu Zhang. 2023.
\newblock \href {http://arxiv.org/abs/2310.08475} {Can we edit multimodal large
  language models?}

\bibitem[{Chowdhery et~al.(2022)Chowdhery, Narang, Devlin, Bosma, Mishra,
  Roberts, Barham, Chung, Sutton, Gehrmann et~al.}]{chowdhery2022palm}
Aakanksha Chowdhery, Sharan Narang, Jacob Devlin, Maarten Bosma, Gaurav Mishra,
  Adam Roberts, Paul Barham, Hyung~Won Chung, Charles Sutton, Sebastian
  Gehrmann, et~al. 2022.
\newblock Palm: Scaling language modeling with pathways.
\newblock \emph{arXiv preprint arXiv:2204.02311}.

\bibitem[{Cohen et~al.(2023)Cohen, Biran, Yoran, Globerson, and
  Geva}]{cohen2023evaluating}
Roi Cohen, Eden Biran, Ori Yoran, Amir Globerson, and Mor Geva. 2023.
\newblock \href {http://arxiv.org/abs/2307.12976} {Evaluating the ripple
  effects of knowledge editing in language models}.

\bibitem[{Dagan et~al.(2005)Dagan, Glickman, and Magnini}]{dagan2005pascal}
Ido Dagan, Oren Glickman, and Bernardo Magnini. 2005.
\newblock The pascal recognising textual entailment challenge.
\newblock In \emph{Machine learning challenges workshop}, pages 177--190.
  Springer.

\bibitem[{Dai et~al.(2022{\natexlab{a}})Dai, Dong, Hao, Sui, Chang, and
  Wei}]{dai2022knowledge}
Damai Dai, Li~Dong, Yaru Hao, Zhifang Sui, Baobao Chang, and Furu Wei.
  2022{\natexlab{a}}.
\newblock Knowledge neurons in pretrained transformers.
\newblock In \emph{Proceedings of the 60th Annual Meeting of the Association
  for Computational Linguistics (Volume 1: Long Papers)}, pages 8493--8502.

\bibitem[{Dai et~al.(2022{\natexlab{b}})Dai, Jiang, Dong, Lyu, She, and
  Sui}]{dai2022neural}
Damai Dai, Wenbin Jiang, Qingxiu Dong, Yajuan Lyu, Qiaoqiao She, and Zhifang
  Sui. 2022{\natexlab{b}}.
\newblock Neural knowledge bank for pretrained transformers.
\newblock \emph{arXiv preprint arXiv:2208.00399}.

\bibitem[{De~Cao et~al.(2021)De~Cao, Aziz, and
  Titov}]{de-cao-etal-2021-editing}
Nicola De~Cao, Wilker Aziz, and Ivan Titov. 2021.
\newblock \href {https://doi.org/10.18653/v1/2021.emnlp-main.522} {Editing
  factual knowledge in language models}.
\newblock In \emph{Proceedings of the 2021 Conference on Empirical Methods in
  Natural Language Processing}, pages 6491--6506, Online and Punta Cana,
  Dominican Republic. Association for Computational Linguistics.

\bibitem[{De~Cao et~al.(2020)De~Cao, Izacard, Riedel, and
  Petroni}]{de2020autoregressive}
Nicola De~Cao, Gautier Izacard, Sebastian Riedel, and Fabio Petroni. 2020.
\newblock Autoregressive entity retrieval.
\newblock \emph{arXiv preprint arXiv:2010.00904}.

\bibitem[{Dong et~al.(2022)Dong, Dai, Song, Xu, Sui, and
  Li}]{dong-etal-2022-calibrating}
Qingxiu Dong, Damai Dai, Yifan Song, Jingjing Xu, Zhifang Sui, and Lei Li.
  2022.
\newblock \href {https://doi.org/10.18653/v1/2022.findings-emnlp.438}
  {Calibrating factual knowledge in pretrained language models}.
\newblock In \emph{Findings of the Association for Computational Linguistics:
  EMNLP 2022}, pages 5937--5947, Abu Dhabi, United Arab Emirates. Association
  for Computational Linguistics.

\bibitem[{Dong et~al.(2023)Dong, Luo, Jiang, Jin, and Li}]{dong2023pace}
Yihong Dong, Kangcheng Luo, Xue Jiang, Zhi Jin, and Ge~Li. 2023.
\newblock Pace: Improving prompt with actor-critic editing for large language
  model.
\newblock \emph{arXiv preprint arXiv:2308.10088}.

\bibitem[{Du et~al.(2022)Du, Bosselut, and Manning}]{du2022synthetic}
Yibing Du, Antoine Bosselut, and Christopher~D Manning. 2022.
\newblock Synthetic disinformation attacks on automated fact verification
  systems.
\newblock In \emph{Proceedings of the AAAI Conference on Artificial
  Intelligence}, volume~36, pages 10581--10589.

\bibitem[{Eisenschlos et~al.(2021)Eisenschlos, Dhingra, Bulian,
  B{\"o}rschinger, and Boyd-Graber}]{eisenschlos-etal-2021-fool}
Julian Eisenschlos, Bhuwan Dhingra, Jannis Bulian, Benjamin B{\"o}rschinger,
  and Jordan Boyd-Graber. 2021.
\newblock \href {https://doi.org/10.18653/v1/2021.naacl-main.32} {Fool me
  twice: Entailment from {W}ikipedia gamification}.
\newblock In \emph{Proceedings of the 2021 Conference of the North American
  Chapter of the Association for Computational Linguistics: Human Language
  Technologies}, pages 352--365, Online. Association for Computational
  Linguistics.

\bibitem[{Feng et~al.(2023{\natexlab{a}})Feng, Shi, Bai, Balachandran, He, and
  Tsvetkov}]{feng2023cook}
Shangbin Feng, Weijia Shi, Yuyang Bai, Vidhisha Balachandran, Tianxing He, and
  Yulia Tsvetkov. 2023{\natexlab{a}}.
\newblock Cook: Empowering general-purpose language models with modular and
  collaborative knowledge.
\newblock \emph{arXiv preprint arXiv:2305.09955}.

\bibitem[{Feng et~al.(2023{\natexlab{b}})Feng, Shi, Bai, Balachandran, He, and
  Tsvetkov}]{feng2023knowledge}
Shangbin Feng, Weijia Shi, Yuyang Bai, Vidhisha Balachandran, Tianxing He, and
  Yulia Tsvetkov. 2023{\natexlab{b}}.
\newblock \href {http://arxiv.org/abs/2305.09955} {Knowledge card: Filling
  llms' knowledge gaps with plug-in specialized language models}.

\bibitem[{Feng et~al.(2023{\natexlab{c}})Feng, Feng, Zhao, Yang, and
  Qin}]{feng2023retrieval}
Zhangyin Feng, Xiaocheng Feng, Dezhi Zhao, Maojin Yang, and Bing Qin.
  2023{\natexlab{c}}.
\newblock Retrieval-generation synergy augmented large language models.
\newblock \emph{arXiv preprint arXiv:2310.05149}.

\bibitem[{Gao et~al.(2023)Gao, Dai, Pasupat, Chen, Chaganty, Fan, Zhao, Lao,
  Lee, Juan, and Guu}]{gao-etal-2023-rarr}
Luyu Gao, Zhuyun Dai, Panupong Pasupat, Anthony Chen, Arun~Tejasvi Chaganty,
  Yicheng Fan, Vincent Zhao, Ni~Lao, Hongrae Lee, Da-Cheng Juan, and Kelvin
  Guu. 2023.
\newblock \href {https://doi.org/10.18653/v1/2023.acl-long.910} {{RARR}:
  Researching and revising what language models say, using language models}.
\newblock In \emph{Proceedings of the 61st Annual Meeting of the Association
  for Computational Linguistics (Volume 1: Long Papers)}, pages 16477--16508,
  Toronto, Canada. Association for Computational Linguistics.

\bibitem[{Geva et~al.(2021{\natexlab{a}})Geva, Khashabi, Segal, Khot, Roth, and
  Berant}]{geva2021did}
Mor Geva, Daniel Khashabi, Elad Segal, Tushar Khot, Dan Roth, and Jonathan
  Berant. 2021{\natexlab{a}}.
\newblock Did aristotle use a laptop? a question answering benchmark with
  implicit reasoning strategies.
\newblock \emph{Transactions of the Association for Computational Linguistics},
  9:346--361.

\bibitem[{Geva et~al.(2021{\natexlab{b}})Geva, Schuster, Berant, and
  Levy}]{geva-etal-2021-transformer}
Mor Geva, Roei Schuster, Jonathan Berant, and Omer Levy. 2021{\natexlab{b}}.
\newblock \href {https://doi.org/10.18653/v1/2021.emnlp-main.446} {Transformer
  feed-forward layers are key-value memories}.
\newblock In \emph{Proceedings of the 2021 Conference on Empirical Methods in
  Natural Language Processing}, pages 5484--5495, Online and Punta Cana,
  Dominican Republic. Association for Computational Linguistics.

\bibitem[{Gu et~al.(2022)Gu, Feng, Ma, Zhang, Gong, and
  Qin}]{gu-etal-2022-distributional}
Yuxuan Gu, Xiaocheng Feng, Sicheng Ma, Lingyuan Zhang, Heng Gong, and Bing Qin.
  2022.
\newblock \href {https://doi.org/10.18653/v1/2022.emnlp-main.67} {A
  distributional lens for multi-aspect controllable text generation}.
\newblock In \emph{Proceedings of the 2022 Conference on Empirical Methods in
  Natural Language Processing}, pages 1023--1043, Abu Dhabi, United Arab
  Emirates. Association for Computational Linguistics.

\bibitem[{Gupta et~al.(2023)Gupta, Mondal, Sheshadri, Zhao, Li, Wiegreffe, and
  Tandon}]{DBLP:journals/corr/abs-2305-14956}
Anshita Gupta, Debanjan Mondal, Akshay~Krishna Sheshadri, Wenlong Zhao,
  Xiang~Lorraine Li, Sarah Wiegreffe, and Niket Tandon. 2023.
\newblock \href {https://doi.org/10.48550/arXiv.2305.14956} {Editing
  commonsense knowledge in {GPT}}.
\newblock \emph{CoRR}, abs/2305.14956.

\bibitem[{Gupta et~al.(2020)Gupta, Mehta, Nokhiz, and
  Srikumar}]{gupta-etal-2020-infotabs}
Vivek Gupta, Maitrey Mehta, Pegah Nokhiz, and Vivek Srikumar. 2020.
\newblock \href {https://doi.org/10.18653/v1/2020.acl-main.210} {{INFOTABS}:
  Inference on tables as semi-structured data}.
\newblock In \emph{Proceedings of the 58th Annual Meeting of the Association
  for Computational Linguistics}, pages 2309--2324, Online. Association for
  Computational Linguistics.

\bibitem[{Hartvigsen et~al.(2022)Hartvigsen, Sankaranarayanan, Palangi, Kim,
  and Ghassemi}]{hartvigsen2022aging}
Thomas Hartvigsen, Swami Sankaranarayanan, Hamid Palangi, Yoon Kim, and Marzyeh
  Ghassemi. 2022.
\newblock Aging with grace: Lifelong model editing with discrete key-value
  adaptors.
\newblock \emph{arXiv preprint arXiv:2211.11031}.

\bibitem[{Hase et~al.(2023)Hase, Bansal, Kim, and
  Ghandeharioun}]{DBLP:journals/corr/abs-2301-04213}
Peter Hase, Mohit Bansal, Been Kim, and Asma Ghandeharioun. 2023.
\newblock \href {https://doi.org/10.48550/arXiv.2301.04213} {Does localization
  inform editing? surprising differences in causality-based localization vs.
  knowledge editing in language models}.
\newblock \emph{CoRR}, abs/2301.04213.

\bibitem[{He et~al.(2022)He, Zhang, and Roth}]{he2022rethinking}
Hangfeng He, Hongming Zhang, and Dan Roth. 2022.
\newblock \href {http://arxiv.org/abs/2301.00303} {Rethinking with retrieval:
  Faithful large language model inference}.

\bibitem[{Ho et~al.(2020)Ho, Duong~Nguyen, Sugawara, and
  Aizawa}]{ho-etal-2020-constructing}
Xanh Ho, Anh-Khoa Duong~Nguyen, Saku Sugawara, and Akiko Aizawa. 2020.
\newblock \href {https://doi.org/10.18653/v1/2020.coling-main.580}
  {Constructing a multi-hop {QA} dataset for comprehensive evaluation of
  reasoning steps}.
\newblock In \emph{Proceedings of the 28th International Conference on
  Computational Linguistics}, pages 6609--6625, Barcelona, Spain (Online).
  International Committee on Computational Linguistics.

\bibitem[{Hoelscher-Obermaier et~al.(2023)Hoelscher-Obermaier, Persson, Kran,
  Konstas, and Barez}]{hoelscherobermaier2023detecting}
Jason Hoelscher-Obermaier, Julia Persson, Esben Kran, Ioannis Konstas, and Fazl
  Barez. 2023.
\newblock \href {http://arxiv.org/abs/2305.17553} {Detecting edit failures in
  large language models: An improved specificity benchmark}.

\bibitem[{Hoffmann et~al.(2022)Hoffmann, Borgeaud, Mensch, Buchatskaya, Cai,
  Rutherford, de~Las~Casas, Hendricks, Welbl, Clark, Hennigan, Noland,
  Millican, van~den Driessche, Damoc, Guy, Osindero, Simonyan, Elsen, Rae,
  Vinyals, and Sifre}]{hoffmann2022training}
Jordan Hoffmann, Sebastian Borgeaud, Arthur Mensch, Elena Buchatskaya, Trevor
  Cai, Eliza Rutherford, Diego de~Las~Casas, Lisa~Anne Hendricks, Johannes
  Welbl, Aidan Clark, Tom Hennigan, Eric Noland, Katie Millican, George van~den
  Driessche, Bogdan Damoc, Aurelia Guy, Simon Osindero, Karen Simonyan, Erich
  Elsen, Jack~W. Rae, Oriol Vinyals, and Laurent Sifre. 2022.
\newblock \href {http://arxiv.org/abs/2203.15556} {Training compute-optimal
  large language models}.

\bibitem[{Hu et~al.(2023)Hu, Liu, Zhao, Hou, Nie, and Li}]{hu2023survey}
Linmei Hu, Zeyi Liu, Ziwang Zhao, Lei Hou, Liqiang Nie, and Juanzi Li. 2023.
\newblock A survey of knowledge enhanced pre-trained language models.
\newblock \emph{IEEE Transactions on Knowledge and Data Engineering}.

\bibitem[{Huang et~al.(2023{\natexlab{a}})Huang, Yu, Ma, Zhong, Feng, Wang,
  Chen, Peng, Feng, Qin, and Liu}]{huang2023survey}
Lei Huang, Weijiang Yu, Weitao Ma, Weihong Zhong, Zhangyin Feng, Haotian Wang,
  Qianglong Chen, Weihua Peng, Xiaocheng Feng, Bing Qin, and Ting Liu.
  2023{\natexlab{a}}.
\newblock \href {http://arxiv.org/abs/2311.05232} {A survey on hallucination in
  large language models: Principles, taxonomy, challenges, and open questions}.

\bibitem[{Huang et~al.(2023{\natexlab{b}})Huang, Shen, Zhang, Zhou, Rong, and
  Xiong}]{huang2023transformer}
Zeyu Huang, Yikang Shen, Xiaofeng Zhang, Jie Zhou, Wenge Rong, and Zhang Xiong.
  2023{\natexlab{b}}.
\newblock Transformer-patcher: One mistake worth one neuron.
\newblock \emph{arXiv preprint arXiv:2301.09785}.

\bibitem[{Ilharco et~al.(2023)Ilharco, Ribeiro, Wortsman, Schmidt, Hajishirzi,
  and Farhadi}]{DBLP:conf/iclr/IlharcoRWSHF23}
Gabriel Ilharco, Marco~T{\'{u}}lio Ribeiro, Mitchell Wortsman, Ludwig Schmidt,
  Hannaneh Hajishirzi, and Ali Farhadi. 2023.
\newblock \href {https://openreview.net/pdf?id=6t0Kwf8-jrj} {Editing models
  with task arithmetic}.
\newblock In \emph{The Eleventh International Conference on Learning
  Representations, {ICLR} 2023, Kigali, Rwanda, May 1-5, 2023}. OpenReview.net.

\bibitem[{Izacard and Grave(2022)}]{izacard2022distilling}
Gautier Izacard and Edouard Grave. 2022.
\newblock \href {http://arxiv.org/abs/2012.04584} {Distilling knowledge from
  reader to retriever for question answering}.

\bibitem[{Izacard et~al.(2022)Izacard, Lewis, Lomeli, Hosseini, Petroni,
  Schick, Dwivedi-Yu, Joulin, Riedel, and Grave}]{izacard2022atlas}
Gautier Izacard, Patrick Lewis, Maria Lomeli, Lucas Hosseini, Fabio Petroni,
  Timo Schick, Jane Dwivedi-Yu, Armand Joulin, Sebastian Riedel, and Edouard
  Grave. 2022.
\newblock \href {http://arxiv.org/abs/2208.03299} {Atlas: Few-shot learning
  with retrieval augmented language models}.

\bibitem[{Jia et~al.(2018)Jia, Abujabal, Saha~Roy, Str{\"o}tgen, and
  Weikum}]{jia2018tempquestions}
Zhen Jia, Abdalghani Abujabal, Rishiraj Saha~Roy, Jannik Str{\"o}tgen, and
  Gerhard Weikum. 2018.
\newblock Tempquestions: A benchmark for temporal question answering.
\newblock In \emph{Companion Proceedings of the The Web Conference 2018}, pages
  1057--1062.

\bibitem[{Jiang et~al.(2021)Jiang, Araki, Ding, and
  Neubig}]{jiang-etal-2021-know}
Zhengbao Jiang, Jun Araki, Haibo Ding, and Graham Neubig. 2021.
\newblock \href {https://doi.org/10.1162/tacl_a_00407} {How can we know when
  language models know? on the calibration of language models for question
  answering}.
\newblock \emph{Transactions of the Association for Computational Linguistics},
  9:962--977.

\bibitem[{Jiang et~al.(2023)Jiang, Xu, Gao, Sun, Liu, Dwivedi-Yu, Yang, Callan,
  and Neubig}]{jiang2023active}
Zhengbao Jiang, Frank~F. Xu, Luyu Gao, Zhiqing Sun, Qian Liu, Jane Dwivedi-Yu,
  Yiming Yang, Jamie Callan, and Graham Neubig. 2023.
\newblock \href {http://arxiv.org/abs/2305.06983} {Active retrieval augmented
  generation}.

\bibitem[{Joshi et~al.(2017)Joshi, Choi, Weld, and
  Zettlemoyer}]{joshi-etal-2017-triviaqa}
Mandar Joshi, Eunsol Choi, Daniel Weld, and Luke Zettlemoyer. 2017.
\newblock \href {https://doi.org/10.18653/v1/P17-1147} {{T}rivia{QA}: A large
  scale distantly supervised challenge dataset for reading comprehension}.
\newblock In \emph{Proceedings of the 55th Annual Meeting of the Association
  for Computational Linguistics (Volume 1: Long Papers)}, pages 1601--1611,
  Vancouver, Canada. Association for Computational Linguistics.

\bibitem[{Ju et~al.(2022)Ju, Yu, Zhao, Zhang, and Ye}]{ju2022grape}
Mingxuan Ju, Wenhao Yu, Tong Zhao, Chuxu Zhang, and Yanfang Ye. 2022.
\newblock \href {http://arxiv.org/abs/2210.02933} {Grape: Knowledge graph
  enhanced passage reader for open-domain question answering}.

\bibitem[{Ju and Zhang(2023)}]{ju2023klob}
Yiming Ju and Zheng Zhang. 2023.
\newblock \href {http://arxiv.org/abs/2309.16535} {Klob: a benchmark for
  assessing knowledge locating methods in language models}.

\bibitem[{Kadavath et~al.(2022)Kadavath, Conerly, Askell, Henighan, Drain,
  Perez, Schiefer, Hatfield-Dodds, DasSarma, Tran-Johnson
  et~al.}]{kadavath2022language}
Saurav Kadavath, Tom Conerly, Amanda Askell, Tom Henighan, Dawn Drain, Ethan
  Perez, Nicholas Schiefer, Zac Hatfield-Dodds, Nova DasSarma, Eli
  Tran-Johnson, et~al. 2022.
\newblock Language models (mostly) know what they know.
\newblock \emph{arXiv preprint arXiv:2207.05221}.

\bibitem[{Kandpal et~al.(2023)Kandpal, Deng, Roberts, Wallace, and
  Raffel}]{kandpal2023large}
Nikhil Kandpal, Haikang Deng, Adam Roberts, Eric Wallace, and Colin Raffel.
  2023.
\newblock Large language models struggle to learn long-tail knowledge.
\newblock In \emph{International Conference on Machine Learning}, pages
  15696--15707. PMLR.

\bibitem[{Karpukhin et~al.(2020)Karpukhin, Oguz, Min, Lewis, Wu, Edunov, Chen,
  and Yih}]{DBLP:conf/emnlp/KarpukhinOMLWEC20}
Vladimir Karpukhin, Barlas Oguz, Sewon Min, Patrick S.~H. Lewis, Ledell Wu,
  Sergey Edunov, Danqi Chen, and Wen{-}tau Yih. 2020.
\newblock \href {https://doi.org/10.18653/v1/2020.emnlp-main.550} {Dense
  passage retrieval for open-domain question answering}.
\newblock In \emph{Proceedings of the 2020 Conference on Empirical Methods in
  Natural Language Processing, {EMNLP} 2020, Online, November 16-20, 2020},
  pages 6769--6781. Association for Computational Linguistics.

\bibitem[{Kasai et~al.(2022)Kasai, Sakaguchi, Takahashi, Bras, Asai, Yu, Radev,
  Smith, Choi, and Inui}]{kasai2022realtime}
Jungo Kasai, Keisuke Sakaguchi, Yoichi Takahashi, Ronan~Le Bras, Akari Asai,
  Xinyan Yu, Dragomir Radev, Noah~A. Smith, Yejin Choi, and Kentaro Inui. 2022.
\newblock \href {http://arxiv.org/abs/2207.13332} {Realtime qa: What's the
  answer right now?}

\bibitem[{Khattab et~al.(2023)Khattab, Santhanam, Li, Hall, Liang, Potts, and
  Zaharia}]{khattab2023demonstratesearchpredict}
Omar Khattab, Keshav Santhanam, Xiang~Lisa Li, David Hall, Percy Liang,
  Christopher Potts, and Matei Zaharia. 2023.
\newblock \href {http://arxiv.org/abs/2212.14024} {Demonstrate-search-predict:
  Composing retrieval and language models for knowledge-intensive nlp}.

\bibitem[{Kwiatkowski et~al.(2019)Kwiatkowski, Palomaki, Redfield, Collins,
  Parikh, Alberti, Epstein, Polosukhin, Devlin, Lee, Toutanova, Jones, Kelcey,
  Chang, Dai, Uszkoreit, Le, and Petrov}]{kwiatkowski-etal-2019-natural}
Tom Kwiatkowski, Jennimaria Palomaki, Olivia Redfield, Michael Collins, Ankur
  Parikh, Chris Alberti, Danielle Epstein, Illia Polosukhin, Jacob Devlin,
  Kenton Lee, Kristina Toutanova, Llion Jones, Matthew Kelcey, Ming-Wei Chang,
  Andrew~M. Dai, Jakob Uszkoreit, Quoc Le, and Slav Petrov. 2019.
\newblock \href {https://doi.org/10.1162/tacl_a_00276} {Natural questions: A
  benchmark for question answering research}.
\newblock \emph{Transactions of the Association for Computational Linguistics},
  7:452--466.

\bibitem[{Lazaridou et~al.(2022)Lazaridou, Gribovskaya, Stokowiec, and
  Grigorev}]{lazaridou2022internet}
Angeliki Lazaridou, Elena Gribovskaya, Wojciech Stokowiec, and Nikolai
  Grigorev. 2022.
\newblock Internet-augmented language models through few-shot prompting for
  open-domain question answering.
\newblock \emph{arXiv preprint arXiv:2203.05115}.

\bibitem[{Lee et~al.(2022)Lee, Yang, Oh, and Seo}]{lee2022generative}
Hyunji Lee, Sohee Yang, Hanseok Oh, and Minjoon Seo. 2022.
\newblock Generative multi-hop retrieval.
\newblock In \emph{Proceedings of the 2022 Conference on Empirical Methods in
  Natural Language Processing}, pages 1417--1436.

\bibitem[{Levy et~al.(2017)Levy, Seo, Choi, and
  Zettlemoyer}]{levy-etal-2017-zero}
Omer Levy, Minjoon Seo, Eunsol Choi, and Luke Zettlemoyer. 2017.
\newblock \href {https://doi.org/10.18653/v1/K17-1034} {Zero-shot relation
  extraction via reading comprehension}.
\newblock In \emph{Proceedings of the 21st Conference on Computational Natural
  Language Learning ({C}o{NLL} 2017)}, pages 333--342, Vancouver, Canada.
  Association for Computational Linguistics.

\bibitem[{Lewis et~al.(2019)Lewis, Liu, Goyal, Ghazvininejad, Mohamed, Levy,
  Stoyanov, and Zettlemoyer}]{lewis2019bart}
Mike Lewis, Yinhan Liu, Naman Goyal, Marjan Ghazvininejad, Abdelrahman Mohamed,
  Omer Levy, Ves Stoyanov, and Luke Zettlemoyer. 2019.
\newblock Bart: Denoising sequence-to-sequence pre-training for natural
  language generation, translation, and comprehension.
\newblock \emph{arXiv preprint arXiv:1910.13461}.

\bibitem[{Lewis et~al.(2021)Lewis, Perez, Piktus, Petroni, Karpukhin, Goyal,
  Küttler, Lewis, tau Yih, Rocktäschel, Riedel, and
  Kiela}]{lewis2021retrievalaugmented}
Patrick Lewis, Ethan Perez, Aleksandra Piktus, Fabio Petroni, Vladimir
  Karpukhin, Naman Goyal, Heinrich Küttler, Mike Lewis, Wen tau Yih, Tim
  Rocktäschel, Sebastian Riedel, and Douwe Kiela. 2021.
\newblock \href {http://arxiv.org/abs/2005.11401} {Retrieval-augmented
  generation for knowledge-intensive nlp tasks}.

\bibitem[{Li et~al.(2022)Li, Rawat, Zaheer, Wang, Lukasik, Veit, Yu, and
  Kumar}]{li2022large}
Daliang Li, Ankit~Singh Rawat, Manzil Zaheer, Xin Wang, Michal Lukasik, Andreas
  Veit, Felix Yu, and Sanjiv Kumar. 2022.
\newblock \href {http://arxiv.org/abs/2211.05110} {Large language models with
  controllable working memory}.

\bibitem[{Li et~al.(2023{\natexlab{a}})Li, Hammoud, Itani, Khizbullin, and
  Ghanem}]{li2023camel}
Guohao Li, Hasan Abed Al~Kader Hammoud, Hani Itani, Dmitrii Khizbullin, and
  Bernard Ghanem. 2023{\natexlab{a}}.
\newblock \href {http://arxiv.org/abs/2303.17760} {Camel: Communicative agents
  for "mind" exploration of large language model society}.

\bibitem[{Li et~al.(2023{\natexlab{b}})Li, Li, Savarese, and Hoi}]{li2023blip}
Junnan Li, Dongxu Li, Silvio Savarese, and Steven Hoi. 2023{\natexlab{b}}.
\newblock Blip-2: Bootstrapping language-image pre-training with frozen image
  encoders and large language models.
\newblock \emph{arXiv preprint arXiv:2301.12597}.

\bibitem[{Li et~al.(2023{\natexlab{c}})Li, Li, Song, Yang, Ma, and
  Yu}]{DBLP:journals/corr/abs-2308-08742}
Xiaopeng Li, Shasha Li, Shezheng Song, Jing Yang, Jun Ma, and Jie Yu.
  2023{\natexlab{c}}.
\newblock \href {https://doi.org/10.48550/arXiv.2308.08742} {{PMET:} precise
  model editing in a transformer}.
\newblock \emph{CoRR}, abs/2308.08742.

\bibitem[{Li et~al.(2023{\natexlab{d}})Li, Yang, Wang, Wei, and
  Li}]{li-etal-2023-multiview}
Yongqi Li, Nan Yang, Liang Wang, Furu Wei, and Wenjie Li. 2023{\natexlab{d}}.
\newblock \href {https://doi.org/10.18653/v1/2023.acl-long.366} {Multiview
  identifiers enhanced generative retrieval}.
\newblock In \emph{Proceedings of the 61st Annual Meeting of the Association
  for Computational Linguistics (Volume 1: Long Papers)}, pages 6636--6648,
  Toronto, Canada. Association for Computational Linguistics.

\bibitem[{Li et~al.(2023{\natexlab{e}})Li, Li, Zhang, Dan, Jiang, and
  Zhang}]{li2023chatdoctor}
Yunxiang Li, Zihan Li, Kai Zhang, Ruilong Dan, Steve Jiang, and You Zhang.
  2023{\natexlab{e}}.
\newblock Chatdoctor: A medical chat model fine-tuned on a large language model
  meta-ai (llama) using medical domain knowledge.
\newblock \emph{Cureus}, 15(6).

\bibitem[{Li et~al.(2023{\natexlab{f}})Li, Zhang, Yao, Wang, Chen, and
  Chen}]{li2023unveiling}
Zhoubo Li, Ningyu Zhang, Yunzhi Yao, Mengru Wang, Xi~Chen, and Huajun Chen.
  2023{\natexlab{f}}.
\newblock \href {http://arxiv.org/abs/2310.02129} {Unveiling the pitfalls of
  knowledge editing for large language models}.

\bibitem[{Liška et~al.(2022)Liška, Kočiský, Gribovskaya, Terzi, Sezener,
  Agrawal, de~Masson~d'Autume, Scholtes, Zaheer, Young, Gilsenan-McMahon,
  Austin, Blunsom, and Lazaridou}]{liska2022streamingqa}
Adam Liška, Tomáš Kočiský, Elena Gribovskaya, Tayfun Terzi, Eren Sezener,
  Devang Agrawal, Cyprien de~Masson~d'Autume, Tim Scholtes, Manzil Zaheer,
  Susannah Young, Ellen Gilsenan-McMahon, Sophia Austin, Phil Blunsom, and
  Angeliki Lazaridou. 2022.
\newblock \href {http://arxiv.org/abs/2205.11388} {Streamingqa: A benchmark for
  adaptation to new knowledge over time in question answering models}.

\bibitem[{Longpre et~al.(2021)Longpre, Perisetla, Chen, Ramesh, DuBois, and
  Singh}]{longpre-etal-2021-entity}
Shayne Longpre, Kartik Perisetla, Anthony Chen, Nikhil Ramesh, Chris DuBois,
  and Sameer Singh. 2021.
\newblock \href {https://doi.org/10.18653/v1/2021.emnlp-main.565} {Entity-based
  knowledge conflicts in question answering}.
\newblock In \emph{Proceedings of the 2021 Conference on Empirical Methods in
  Natural Language Processing}, pages 7052--7063, Online and Punta Cana,
  Dominican Republic. Association for Computational Linguistics.

\bibitem[{Luo et~al.(2023)Luo, Xu, Zhao, Geng, Tao, Ma, Lin, and
  Jiang}]{luo2023augmented}
Ziyang Luo, Can Xu, Pu~Zhao, Xiubo Geng, Chongyang Tao, Jing Ma, Qingwei Lin,
  and Daxin Jiang. 2023.
\newblock Augmented large language models with parametric knowledge guiding.
\newblock \emph{arXiv preprint arXiv:2305.04757}.

\bibitem[{Ma et~al.(2023)Ma, Gong, He, Zhao, and Duan}]{ma2023query}
Xinbei Ma, Yeyun Gong, Pengcheng He, Hai Zhao, and Nan Duan. 2023.
\newblock Query rewriting for retrieval-augmented large language models.
\newblock \emph{arXiv preprint arXiv:2305.14283}.

\bibitem[{Madaan et~al.(2022)Madaan, Tandon, Clark, and
  Yang}]{madaan-etal-2022-memory}
Aman Madaan, Niket Tandon, Peter Clark, and Yiming Yang. 2022.
\newblock \href {https://doi.org/10.18653/v1/2022.emnlp-main.183}
  {Memory-assisted prompt editing to improve {GPT}-3 after deployment}.
\newblock In \emph{Proceedings of the 2022 Conference on Empirical Methods in
  Natural Language Processing}, pages 2833--2861, Abu Dhabi, United Arab
  Emirates. Association for Computational Linguistics.

\bibitem[{Mallen et~al.(2023)Mallen, Asai, Zhong, Das, Khashabi, and
  Hajishirzi}]{mallen-etal-2023-trust}
Alex Mallen, Akari Asai, Victor Zhong, Rajarshi Das, Daniel Khashabi, and
  Hannaneh Hajishirzi. 2023.
\newblock \href {https://doi.org/10.18653/v1/2023.acl-long.546} {When not to
  trust language models: Investigating effectiveness of parametric and
  non-parametric memories}.
\newblock In \emph{Proceedings of the 61st Annual Meeting of the Association
  for Computational Linguistics (Volume 1: Long Papers)}, pages 9802--9822,
  Toronto, Canada. Association for Computational Linguistics.

\bibitem[{Manakul et~al.(2023)Manakul, Liusie, and
  Gales}]{manakul2023selfcheckgpt}
Potsawee Manakul, Adian Liusie, and Mark J.~F. Gales. 2023.
\newblock \href {http://arxiv.org/abs/2303.08896} {Selfcheckgpt: Zero-resource
  black-box hallucination detection for generative large language models}.

\bibitem[{Mao et~al.(2023)Mao, Zhang, Wang, Wang, Yao, Jiang, Xie, Huang, and
  Chen}]{mao2023editing}
Shengyu Mao, Ningyu Zhang, Xiaohan Wang, Mengru Wang, Yunzhi Yao, Yong Jiang,
  Pengjun Xie, Fei Huang, and Huajun Chen. 2023.
\newblock \href {http://arxiv.org/abs/2310.02168} {Editing personality for
  llms}.

\bibitem[{Maronikolakis and
  Sch{\"u}tze(2021)}]{maronikolakis-schutze-2021-multidomain}
Antonis Maronikolakis and Hinrich Sch{\"u}tze. 2021.
\newblock \href {https://aclanthology.org/2021.adaptnlp-1.1} {Multidomain
  pretrained language models for green {NLP}}.
\newblock In \emph{Proceedings of the Second Workshop on Domain Adaptation for
  NLP}, pages 1--8, Kyiv, Ukraine. Association for Computational Linguistics.

\bibitem[{Meng et~al.(2022)Meng, Bau, Andonian, and
  Belinkov}]{DBLP:conf/nips/MengBAB22}
Kevin Meng, David Bau, Alex Andonian, and Yonatan Belinkov. 2022.
\newblock \href
  {http://papers.nips.cc/paper\_files/paper/2022/hash/6f1d43d5a82a37e89b0665b33bf3a182-Abstract-Conference.html}
  {Locating and editing factual associations in {GPT}}.
\newblock In \emph{NeurIPS}.

\bibitem[{Meng et~al.(2023)Meng, Sharma, Andonian, Belinkov, and
  Bau}]{DBLP:conf/iclr/MengSABB23}
Kevin Meng, Arnab~Sen Sharma, Alex~J. Andonian, Yonatan Belinkov, and David
  Bau. 2023.
\newblock \href {https://openreview.net/pdf?id=MkbcAHIYgyS} {Mass-editing
  memory in a transformer}.
\newblock In \emph{The Eleventh International Conference on Learning
  Representations, {ICLR} 2023, Kigali, Rwanda, May 1-5, 2023}. OpenReview.net.

\bibitem[{Mitchell et~al.(2022{\natexlab{a}})Mitchell, Lin, Bosselut, Finn, and
  Manning}]{DBLP:conf/iclr/MitchellLBFM22}
Eric Mitchell, Charles Lin, Antoine Bosselut, Chelsea Finn, and Christopher~D.
  Manning. 2022{\natexlab{a}}.
\newblock \href {https://openreview.net/forum?id=0DcZxeWfOPt} {Fast model
  editing at scale}.
\newblock In \emph{The Tenth International Conference on Learning
  Representations, {ICLR} 2022, Virtual Event, April 25-29, 2022}.
  OpenReview.net.

\bibitem[{Mitchell et~al.(2022{\natexlab{b}})Mitchell, Lin, Bosselut, Manning,
  and Finn}]{mitchell2022memory}
Eric Mitchell, Charles Lin, Antoine Bosselut, Christopher~D Manning, and
  Chelsea Finn. 2022{\natexlab{b}}.
\newblock Memory-based model editing at scale.
\newblock In \emph{International Conference on Machine Learning}, pages
  15817--15831. PMLR.

\bibitem[{Moiseev et~al.(2022)Moiseev, Dong, Alfonseca, and
  Jaggi}]{moiseev-etal-2022-skill}
Fedor Moiseev, Zhe Dong, Enrique Alfonseca, and Martin Jaggi. 2022.
\newblock \href {https://doi.org/10.18653/v1/2022.naacl-main.113} {{SKILL}:
  Structured knowledge infusion for large language models}.
\newblock In \emph{Proceedings of the 2022 Conference of the North American
  Chapter of the Association for Computational Linguistics: Human Language
  Technologies}, pages 1581--1588, Seattle, United States. Association for
  Computational Linguistics.

\bibitem[{Neeman et~al.(2022)Neeman, Aharoni, Honovich, Choshen, Szpektor, and
  Abend}]{neeman2022disentqa}
Ella Neeman, Roee Aharoni, Or~Honovich, Leshem Choshen, Idan Szpektor, and Omri
  Abend. 2022.
\newblock \href {http://arxiv.org/abs/2211.05655} {Disentqa: Disentangling
  parametric and contextual knowledge with counterfactual question answering}.

\bibitem[{OpenAI(2023)}]{openai2023gpt4}
OpenAI. 2023.
\newblock \href {http://arxiv.org/abs/2303.08774} {Gpt-4 technical report}.

\bibitem[{Ouyang et~al.(2022)Ouyang, Wu, Jiang, Almeida, Wainwright, Mishkin,
  Zhang, Agarwal, Slama, Ray, Schulman, Hilton, Kelton, Miller, Simens, Askell,
  Welinder, Christiano, Leike, and Lowe}]{ouyang2022training}
Long Ouyang, Jeff Wu, Xu~Jiang, Diogo Almeida, Carroll~L. Wainwright, Pamela
  Mishkin, Chong Zhang, Sandhini Agarwal, Katarina Slama, Alex Ray, John
  Schulman, Jacob Hilton, Fraser Kelton, Luke Miller, Maddie Simens, Amanda
  Askell, Peter Welinder, Paul Christiano, Jan Leike, and Ryan Lowe. 2022.
\newblock \href {http://arxiv.org/abs/2203.02155} {Training language models to
  follow instructions with human feedback}.

\bibitem[{Padgham and Winikoff(2005)}]{padgham2005developing}
Lin Padgham and Michael Winikoff. 2005.
\newblock \emph{Developing intelligent agent systems: A practical guide}.
\newblock John Wiley \& Sons.

\bibitem[{Patil et~al.(2023)Patil, Hase, and Bansal}]{patil2023sensitive}
Vaidehi Patil, Peter Hase, and Mohit Bansal. 2023.
\newblock \href {http://arxiv.org/abs/2309.17410} {Can sensitive information be
  deleted from llms? objectives for defending against extraction attacks}.

\bibitem[{Peng et~al.(2023)Peng, Galley, He, Cheng, Xie, Hu, Huang, Liden, Yu,
  Chen, and Gao}]{peng2023check}
Baolin Peng, Michel Galley, Pengcheng He, Hao Cheng, Yujia Xie, Yu~Hu, Qiuyuan
  Huang, Lars Liden, Zhou Yu, Weizhu Chen, and Jianfeng Gao. 2023.
\newblock \href {http://arxiv.org/abs/2302.12813} {Check your facts and try
  again: Improving large language models with external knowledge and automated
  feedback}.

\bibitem[{Petroni et~al.(2021)Petroni, Piktus, Fan, Lewis, Yazdani, De~Cao,
  Thorne, Jernite, Karpukhin, Maillard, Plachouras, Rockt{\"a}schel, and
  Riedel}]{petroni-etal-2021-kilt}
Fabio Petroni, Aleksandra Piktus, Angela Fan, Patrick Lewis, Majid Yazdani,
  Nicola De~Cao, James Thorne, Yacine Jernite, Vladimir Karpukhin, Jean
  Maillard, Vassilis Plachouras, Tim Rockt{\"a}schel, and Sebastian Riedel.
  2021.
\newblock \href {https://doi.org/10.18653/v1/2021.naacl-main.200} {{KILT}: a
  benchmark for knowledge intensive language tasks}.
\newblock In \emph{Proceedings of the 2021 Conference of the North American
  Chapter of the Association for Computational Linguistics: Human Language
  Technologies}, pages 2523--2544, Online. Association for Computational
  Linguistics.

\bibitem[{Pinter and Elhadad(2023)}]{pinter2023emptying}
Yuval Pinter and Michael Elhadad. 2023.
\newblock \href {http://arxiv.org/abs/2310.11958} {Emptying the ocean with a
  spoon: Should we edit models?}

\bibitem[{Press et~al.(2023)Press, Zhang, Min, Schmidt, Smith, and
  Lewis}]{press2023measuring}
Ofir Press, Muru Zhang, Sewon Min, Ludwig Schmidt, Noah~A. Smith, and Mike
  Lewis. 2023.
\newblock \href {http://arxiv.org/abs/2210.03350} {Measuring and narrowing the
  compositionality gap in language models}.

\bibitem[{Qian et~al.(2023)Qian, Cong, Liu, Yang, Chen, Su, Dang, Li, Xu, Li,
  Liu, and Sun}]{qian2023communicative}
Chen Qian, Xin Cong, Wei Liu, Cheng Yang, Weize Chen, Yusheng Su, Yufan Dang,
  Jiahao Li, Juyuan Xu, Dahai Li, Zhiyuan Liu, and Maosong Sun. 2023.
\newblock \href {http://arxiv.org/abs/2307.07924} {Communicative agents for
  software development}.

\bibitem[{Qian et~al.(2022)Qian, Dong, Shen, Wei, and
  Chen}]{qian-etal-2022-controllable}
Jing Qian, Li~Dong, Yelong Shen, Furu Wei, and Weizhu Chen. 2022.
\newblock \href {https://doi.org/10.18653/v1/2022.findings-acl.229}
  {Controllable natural language generation with contrastive prefixes}.
\newblock In \emph{Findings of the Association for Computational Linguistics:
  ACL 2022}, pages 2912--2924, Dublin, Ireland. Association for Computational
  Linguistics.

\bibitem[{Qiu et~al.(2023)Qiu, Zhang, Li, He, and Lan}]{qiu2023latent}
Huachuan Qiu, Shuai Zhang, Anqi Li, Hongliang He, and Zhenzhong Lan. 2023.
\newblock \href {http://arxiv.org/abs/2307.08487} {Latent jailbreak: A
  benchmark for evaluating text safety and output robustness of large language
  models}.

\bibitem[{Ram et~al.(2023)Ram, Levine, Dalmedigos, Muhlgay, Shashua,
  Leyton-Brown, and Shoham}]{ram2023incontext}
Ori Ram, Yoav Levine, Itay Dalmedigos, Dor Muhlgay, Amnon Shashua, Kevin
  Leyton-Brown, and Yoav Shoham. 2023.
\newblock \href {http://arxiv.org/abs/2302.00083} {In-context
  retrieval-augmented language models}.

\bibitem[{Rawte et~al.(2023)Rawte, Sheth, and Das}]{rawte2023survey}
Vipula Rawte, Amit Sheth, and Amitava Das. 2023.
\newblock \href {http://arxiv.org/abs/2309.05922} {A survey of hallucination in
  large foundation models}.

\bibitem[{Ren et~al.(2023)Ren, Wang, Qu, Zhao, Liu, Tian, Wu, Wen, and
  Wang}]{ren2023investigating}
Ruiyang Ren, Yuhao Wang, Yingqi Qu, Wayne~Xin Zhao, Jing Liu, Hao Tian, Hua Wu,
  Ji-Rong Wen, and Haifeng Wang. 2023.
\newblock Investigating the factual knowledge boundary of large language models
  with retrieval augmentation.
\newblock \emph{arXiv preprint arXiv:2307.11019}.

\bibitem[{Robertson et~al.(2009)Robertson, Zaragoza
  et~al.}]{robertson2009probabilistic}
Stephen Robertson, Hugo Zaragoza, et~al. 2009.
\newblock The probabilistic relevance framework: Bm25 and beyond.
\newblock \emph{Foundations and Trends{\textregistered} in Information
  Retrieval}, 3(4):333--389.

\bibitem[{Saha et~al.(2018)Saha, Pahuja, Khapra, Sankaranarayanan, and
  Chandar}]{saha2018complex}
Amrita Saha, Vardaan Pahuja, Mitesh Khapra, Karthik Sankaranarayanan, and
  Sarath Chandar. 2018.
\newblock Complex sequential question answering: Towards learning to converse
  over linked question answer pairs with a knowledge graph.
\newblock In \emph{Proceedings of the AAAI conference on artificial
  intelligence}, volume~32.

\bibitem[{Sap et~al.(2020)Sap, Shwartz, Bosselut, Choi, and
  Roth}]{sap-etal-2020-commonsense}
Maarten Sap, Vered Shwartz, Antoine Bosselut, Yejin Choi, and Dan Roth. 2020.
\newblock \href {https://doi.org/10.18653/v1/2020.acl-tutorials.7} {Commonsense
  reasoning for natural language processing}.
\newblock In \emph{Proceedings of the 58th Annual Meeting of the Association
  for Computational Linguistics: Tutorial Abstracts}, pages 27--33, Online.
  Association for Computational Linguistics.

\bibitem[{Shao et~al.(2023)Shao, Gong, Shen, Huang, Duan, and
  Chen}]{shao2023enhancing}
Zhihong Shao, Yeyun Gong, Yelong Shen, Minlie Huang, Nan Duan, and Weizhu Chen.
  2023.
\newblock \href {http://arxiv.org/abs/2305.15294} {Enhancing
  retrieval-augmented large language models with iterative retrieval-generation
  synergy}.

\bibitem[{Shi et~al.(2023)Shi, Min, Yasunaga, Seo, James, Lewis, Zettlemoyer,
  and tau Yih}]{shi2023replug}
Weijia Shi, Sewon Min, Michihiro Yasunaga, Minjoon Seo, Rich James, Mike Lewis,
  Luke Zettlemoyer, and Wen tau Yih. 2023.
\newblock \href {http://arxiv.org/abs/2301.12652} {Replug: Retrieval-augmented
  black-box language models}.

\bibitem[{Si et~al.(2022)Si, Gan, Yang, Wang, Wang, Boyd-Graber, and
  Wang}]{si2022prompting}
Chenglei Si, Zhe Gan, Zhengyuan Yang, Shuohang Wang, Jianfeng Wang, Jordan
  Boyd-Graber, and Lijuan Wang. 2022.
\newblock Prompting gpt-3 to be reliable.
\newblock \emph{arXiv preprint arXiv:2210.09150}.

\bibitem[{Sinitsin et~al.(2020)Sinitsin, Plokhotnyuk, Pyrkin, Popov, and
  Babenko}]{sinitsin2020editable}
Anton Sinitsin, Vsevolod Plokhotnyuk, Dmitriy Pyrkin, Sergei Popov, and Artem
  Babenko. 2020.
\newblock Editable neural networks.
\newblock \emph{arXiv preprint arXiv:2004.00345}.

\bibitem[{Sumers et~al.(2023)Sumers, Yao, Narasimhan, and
  Griffiths}]{sumers2023cognitive}
Theodore~R. Sumers, Shunyu Yao, Karthik Narasimhan, and Thomas~L. Griffiths.
  2023.
\newblock \href {http://arxiv.org/abs/2309.02427} {Cognitive architectures for
  language agents}.

\bibitem[{Sun et~al.(2023)Sun, Wang, Tay, Yang, and
  Zhou}]{sun2023recitationaugmented}
Zhiqing Sun, Xuezhi Wang, Yi~Tay, Yiming Yang, and Denny Zhou. 2023.
\newblock \href {http://arxiv.org/abs/2210.01296} {Recitation-augmented
  language models}.

\bibitem[{Talmor et~al.(2019)Talmor, Herzig, Lourie, and
  Berant}]{talmor-etal-2019-commonsenseqa}
Alon Talmor, Jonathan Herzig, Nicholas Lourie, and Jonathan Berant. 2019.
\newblock \href {https://doi.org/10.18653/v1/N19-1421} {{C}ommonsense{QA}: A
  question answering challenge targeting commonsense knowledge}.
\newblock In \emph{Proceedings of the 2019 Conference of the North {A}merican
  Chapter of the Association for Computational Linguistics: Human Language
  Technologies, Volume 1 (Long and Short Papers)}, pages 4149--4158,
  Minneapolis, Minnesota. Association for Computational Linguistics.

\bibitem[{Talmor et~al.(2022)Talmor, Yoran, Bras, Bhagavatula, Goldberg, Choi,
  and Berant}]{talmor2022commonsenseqa}
Alon Talmor, Ori Yoran, Ronan~Le Bras, Chandra Bhagavatula, Yoav Goldberg,
  Yejin Choi, and Jonathan Berant. 2022.
\newblock Commonsenseqa 2.0: Exposing the limits of ai through gamification.
\newblock \emph{arXiv preprint arXiv:2201.05320}.

\bibitem[{Tay et~al.(2022)Tay, Tran, Dehghani, Ni, Bahri, Mehta, Qin, Hui,
  Zhao, Gupta et~al.}]{tay2022transformer}
Yi~Tay, Vinh Tran, Mostafa Dehghani, Jianmo Ni, Dara Bahri, Harsh Mehta, Zhen
  Qin, Kai Hui, Zhe Zhao, Jai Gupta, et~al. 2022.
\newblock Transformer memory as a differentiable search index.
\newblock \emph{Advances in Neural Information Processing Systems},
  35:21831--21843.

\bibitem[{Thorne et~al.(2018)Thorne, Vlachos, Christodoulopoulos, and
  Mittal}]{thorne-etal-2018-fever}
James Thorne, Andreas Vlachos, Christos Christodoulopoulos, and Arpit Mittal.
  2018.
\newblock \href {https://doi.org/10.18653/v1/N18-1074} {{FEVER}: a large-scale
  dataset for fact extraction and {VER}ification}.
\newblock In \emph{Proceedings of the 2018 Conference of the North {A}merican
  Chapter of the Association for Computational Linguistics: Human Language
  Technologies, Volume 1 (Long Papers)}, pages 809--819, New Orleans,
  Louisiana. Association for Computational Linguistics.

\bibitem[{Touvron et~al.(2023)Touvron, Lavril, Izacard, Martinet, Lachaux,
  Lacroix, Rozière, Goyal, Hambro, Azhar, Rodriguez, Joulin, Grave, and
  Lample}]{touvron2023llama}
Hugo Touvron, Thibaut Lavril, Gautier Izacard, Xavier Martinet, Marie-Anne
  Lachaux, Timothée Lacroix, Baptiste Rozière, Naman Goyal, Eric Hambro,
  Faisal Azhar, Aurelien Rodriguez, Armand Joulin, Edouard Grave, and Guillaume
  Lample. 2023.
\newblock \href {http://arxiv.org/abs/2302.13971} {Llama: Open and efficient
  foundation language models}.

\bibitem[{Trivedi et~al.(2022)Trivedi, Balasubramanian, Khot, and
  Sabharwal}]{trivedi2022musique}
Harsh Trivedi, Niranjan Balasubramanian, Tushar Khot, and Ashish Sabharwal.
  2022.
\newblock \href {http://arxiv.org/abs/2108.00573} {Musique: Multihop questions
  via single-hop question composition}.

\bibitem[{Trivedi et~al.(2023)Trivedi, Balasubramanian, Khot, and
  Sabharwal}]{trivedi2023interleaving}
Harsh Trivedi, Niranjan Balasubramanian, Tushar Khot, and Ashish Sabharwal.
  2023.
\newblock \href {http://arxiv.org/abs/2212.10509} {Interleaving retrieval with
  chain-of-thought reasoning for knowledge-intensive multi-step questions}.

\bibitem[{Vu et~al.(2023)Vu, Iyyer, Wang, Constant, Wei, Wei, Tar, Sung, Zhou,
  Le, and Luong}]{vu2023freshllms}
Tu~Vu, Mohit Iyyer, Xuezhi Wang, Noah Constant, Jerry Wei, Jason Wei, Chris
  Tar, Yun-Hsuan Sung, Denny Zhou, Quoc Le, and Thang Luong. 2023.
\newblock \href {http://arxiv.org/abs/2310.03214} {Freshllms: Refreshing large
  language models with search engine augmentation}.

\bibitem[{Wang et~al.(2023{\natexlab{a}})Wang, Liang, Sun, Cao, and
  Xu}]{wang2023crosslingual}
Jiaan Wang, Yunlong Liang, Zengkui Sun, Yuxuan Cao, and Jiarong Xu.
  2023{\natexlab{a}}.
\newblock \href {http://arxiv.org/abs/2309.08952} {Cross-lingual knowledge
  editing in large language models}.

\bibitem[{Wang et~al.(2023{\natexlab{b}})Wang, Yang, and
  Wei}]{wang2023query2doc}
Liang Wang, Nan Yang, and Furu Wei. 2023{\natexlab{b}}.
\newblock \href {http://arxiv.org/abs/2303.07678} {Query2doc: Query expansion
  with large language models}.

\bibitem[{Wang et~al.(2022)Wang, Hou, Wang, Miao, Wu, Chen, Xia, Chi, Zhao, Liu
  et~al.}]{wang2022neural}
Yujing Wang, Yingyan Hou, Haonan Wang, Ziming Miao, Shibin Wu, Qi~Chen, Yuqing
  Xia, Chengmin Chi, Guoshuai Zhao, Zheng Liu, et~al. 2022.
\newblock A neural corpus indexer for document retrieval.
\newblock \emph{Advances in Neural Information Processing Systems},
  35:25600--25614.

\bibitem[{Wei et~al.(2022)Wei, Wang, Schuurmans, Bosma, Xia, Chi, Le, Zhou
  et~al.}]{wei2022chain}
Jason Wei, Xuezhi Wang, Dale Schuurmans, Maarten Bosma, Fei Xia, Ed~Chi, Quoc~V
  Le, Denny Zhou, et~al. 2022.
\newblock Chain-of-thought prompting elicits reasoning in large language
  models.
\newblock \emph{Advances in Neural Information Processing Systems},
  35:24824--24837.

\bibitem[{Wu et~al.(2023)Wu, Peng, Chen, Su, and
  Sun}]{DBLP:journals/corr/abs-2308-09954}
Suhang Wu, Minlong Peng, Yue Chen, Jinsong Su, and Mingming Sun. 2023.
\newblock \href {https://doi.org/10.48550/arXiv.2308.09954} {Eva-kellm: {A} new
  benchmark for evaluating knowledge editing of llms}.
\newblock \emph{CoRR}, abs/2308.09954.

\bibitem[{Xie et~al.(2023)Xie, Zhang, Chen, Lou, and Su}]{xie2023adaptive}
Jian Xie, Kai Zhang, Jiangjie Chen, Renze Lou, and Yu~Su. 2023.
\newblock \href {http://arxiv.org/abs/2305.13300} {Adaptive chameleon or
  stubborn sloth: Unraveling the behavior of large language models in knowledge
  clashes}.

\bibitem[{Xu et~al.(2023)Xu, Xu, Wang, Liu, Zhu, and McAuley}]{xu2023small}
Canwen Xu, Yichong Xu, Shuohang Wang, Yang Liu, Chenguang Zhu, and Julian
  McAuley. 2023.
\newblock Small models are valuable plug-ins for large language models.
\newblock \emph{arXiv preprint arXiv:2305.08848}.

\bibitem[{Yang et~al.(2023)Yang, Chen, Li, Ding, and Wu}]{yang2023chatgpt}
Linyao Yang, Hongyang Chen, Zhao Li, Xiao Ding, and Xindong Wu. 2023.
\newblock Chatgpt is not enough: Enhancing large language models with knowledge
  graphs for fact-aware language modeling.
\newblock \emph{arXiv preprint arXiv:2306.11489}.

\bibitem[{Yang et~al.(2018)Yang, Qi, Zhang, Bengio, Cohen, Salakhutdinov, and
  Manning}]{yang-etal-2018-hotpotqa}
Zhilin Yang, Peng Qi, Saizheng Zhang, Yoshua Bengio, William Cohen, Ruslan
  Salakhutdinov, and Christopher~D. Manning. 2018.
\newblock \href {https://doi.org/10.18653/v1/D18-1259} {{H}otpot{QA}: A dataset
  for diverse, explainable multi-hop question answering}.
\newblock In \emph{Proceedings of the 2018 Conference on Empirical Methods in
  Natural Language Processing}, pages 2369--2380, Brussels, Belgium.
  Association for Computational Linguistics.

\bibitem[{Yao et~al.(2023{\natexlab{a}})Yao, Zhao, Yu, Du, Shafran, Narasimhan,
  and Cao}]{yao2023react}
Shunyu Yao, Jeffrey Zhao, Dian Yu, Nan Du, Izhak Shafran, Karthik Narasimhan,
  and Yuan Cao. 2023{\natexlab{a}}.
\newblock \href {http://arxiv.org/abs/2210.03629} {React: Synergizing reasoning
  and acting in language models}.

\bibitem[{Yao et~al.(2023{\natexlab{b}})Yao, Wang, Tian, Cheng, Li, Deng, Chen,
  and Zhang}]{yao2023editing}
Yunzhi Yao, Peng Wang, Bozhong Tian, Siyuan Cheng, Zhoubo Li, Shumin Deng,
  Huajun Chen, and Ningyu Zhang. 2023{\natexlab{b}}.
\newblock \href {http://arxiv.org/abs/2305.13172} {Editing large language
  models: Problems, methods, and opportunities}.

\bibitem[{Yasunaga et~al.(2023)Yasunaga, Aghajanyan, Shi, James, Leskovec,
  Liang, Lewis, Zettlemoyer, and tau Yih}]{yasunaga2023retrievalaugmented}
Michihiro Yasunaga, Armen Aghajanyan, Weijia Shi, Rich James, Jure Leskovec,
  Percy Liang, Mike Lewis, Luke Zettlemoyer, and Wen tau Yih. 2023.
\newblock \href {http://arxiv.org/abs/2211.12561} {Retrieval-augmented
  multimodal language modeling}.

\bibitem[{Yin et~al.(2022)Yin, Dong, Cheng, Liu, Chang, Wei, and
  Gao}]{yin2022survey}
Da~Yin, Li~Dong, Hao Cheng, Xiaodong Liu, Kai-Wei Chang, Furu Wei, and Jianfeng
  Gao. 2022.
\newblock \href {http://arxiv.org/abs/2202.08772} {A survey of
  knowledge-intensive nlp with pre-trained language models}.

\bibitem[{Yin et~al.(2023)Yin, Sun, Guo, Wu, Qiu, and
  Huang}]{yin-etal-2023-large}
Zhangyue Yin, Qiushi Sun, Qipeng Guo, Jiawen Wu, Xipeng Qiu, and Xuanjing
  Huang. 2023.
\newblock \href {https://doi.org/10.18653/v1/2023.findings-acl.551} {Do large
  language models know what they don{'}t know?}
\newblock In \emph{Findings of the Association for Computational Linguistics:
  ACL 2023}, pages 8653--8665, Toronto, Canada. Association for Computational
  Linguistics.

\bibitem[{Yoran et~al.(2023)Yoran, Wolfson, Ram, and
  Berant}]{Yoran_Wolfson_Ram_Berant_2023}
Ori Yoran, Tomer Wolfson, Ori Ram, and Jonathan Berant. 2023.
\newblock Making retrieval-augmented language models robust to irrelevant
  context.

\bibitem[{Yu et~al.(2021{\natexlab{a}})Yu, Liang, Ji, Li, Fang, Xiao, and
  Duan}]{yu2021hybrid}
Weijiang Yu, Jian Liang, Lei Ji, Lu~Li, Yuejian Fang, Nong Xiao, and Nan Duan.
  2021{\natexlab{a}}.
\newblock Hybrid reasoning network for video-based commonsense captioning.
\newblock In \emph{Proceedings of the 29th ACM international conference on
  multimedia}, pages 5213--5221.

\bibitem[{Yu et~al.(2023{\natexlab{a}})Yu, Wang, Li, Xiao, and
  Ghanem}]{yu2023knowledge}
Weijiang Yu, Haofan Wang, Guohao Li, Nong Xiao, and Bernard Ghanem.
  2023{\natexlab{a}}.
\newblock Knowledge-aware global reasoning for situation recognition.
\newblock \emph{IEEE Transactions on Pattern Analysis and Machine
  Intelligence}.

\bibitem[{Yu et~al.(2021{\natexlab{b}})Yu, Zheng, Li, Ji, Wu, Xiao, and
  Duan}]{yu2021learning}
Weijiang Yu, Haoteng Zheng, Mengfei Li, Lei Ji, Lijun Wu, Nong Xiao, and Nan
  Duan. 2021{\natexlab{b}}.
\newblock Learning from inside: Self-driven siamese sampling and reasoning for
  video question answering.
\newblock \emph{Advances in Neural Information Processing Systems},
  34:26462--26474.

\bibitem[{Yu et~al.(2023{\natexlab{b}})Yu, Iter, Wang, Xu, Ju, Sanyal, Zhu,
  Zeng, and Jiang}]{yu2023generate}
Wenhao Yu, Dan Iter, Shuohang Wang, Yichong Xu, Mingxuan Ju, Soumya Sanyal,
  Chenguang Zhu, Michael Zeng, and Meng Jiang. 2023{\natexlab{b}}.
\newblock \href {http://arxiv.org/abs/2209.10063} {Generate rather than
  retrieve: Large language models are strong context generators}.

\bibitem[{Yu et~al.(2023{\natexlab{c}})Yu, Zhang, Liang, Jiang, and
  Sabharwal}]{yu2023improving}
Wenhao Yu, Zhihan Zhang, Zhenwen Liang, Meng Jiang, and Ashish Sabharwal.
  2023{\natexlab{c}}.
\newblock \href {http://arxiv.org/abs/2305.14002} {Improving language models
  via plug-and-play retrieval feedback}.

\bibitem[{Yu et~al.(2022)Yu, Zhu, Li, Hu, Wang, Ji, and Jiang}]{Yu_2022}
Wenhao Yu, Chenguang Zhu, Zaitang Li, Zhiting Hu, Qingyun Wang, Heng Ji, and
  Meng Jiang. 2022.
\newblock \href {https://doi.org/10.1145/3512467} {A survey of
  knowledge-enhanced text generation}.
\newblock \emph{{ACM} Computing Surveys}, 54(11s):1--38.

\bibitem[{Yu et~al.(2023{\natexlab{d}})Yu, Xiong, Yu, and
  Liu}]{yu-etal-2023-augmentation}
Zichun Yu, Chenyan Xiong, Shi Yu, and Zhiyuan Liu. 2023{\natexlab{d}}.
\newblock \href {https://doi.org/10.18653/v1/2023.acl-long.136}
  {Augmentation-adapted retriever improves generalization of language models as
  generic plug-in}.
\newblock In \emph{Proceedings of the 61st Annual Meeting of the Association
  for Computational Linguistics (Volume 1: Long Papers)}, pages 2421--2436,
  Toronto, Canada. Association for Computational Linguistics.

\bibitem[{Zeng et~al.(2022)Zeng, Liu, Du, Wang, Lai, Ding, Yang, Xu, Zheng, Xia
  et~al.}]{zeng2022glm}
Aohan Zeng, Xiao Liu, Zhengxiao Du, Zihan Wang, Hanyu Lai, Ming Ding, Zhuoyi
  Yang, Yifan Xu, Wendi Zheng, Xiao Xia, et~al. 2022.
\newblock Glm-130b: An open bilingual pre-trained model.
\newblock \emph{arXiv preprint arXiv:2210.02414}.

\bibitem[{Zhang and Choi(2021)}]{zhang-choi-2021-situatedqa}
Michael Zhang and Eunsol Choi. 2021.
\newblock \href {https://doi.org/10.18653/v1/2021.emnlp-main.586}
  {{S}ituated{QA}: Incorporating extra-linguistic contexts into {QA}}.
\newblock In \emph{Proceedings of the 2021 Conference on Empirical Methods in
  Natural Language Processing}, pages 7371--7387, Online and Punta Cana,
  Dominican Republic. Association for Computational Linguistics.

\bibitem[{Zhang et~al.(2023{\natexlab{a}})Zhang, Li, Cui, Cai, Liu, Fu, Huang,
  Zhao, Zhang, Chen, Wang, Luu, Bi, Shi, and Shi}]{zhang2023sirens}
Yue Zhang, Yafu Li, Leyang Cui, Deng Cai, Lemao Liu, Tingchen Fu, Xinting
  Huang, Enbo Zhao, Yu~Zhang, Yulong Chen, Longyue Wang, Anh~Tuan Luu, Wei Bi,
  Freda Shi, and Shuming Shi. 2023{\natexlab{a}}.
\newblock \href {http://arxiv.org/abs/2309.01219} {Siren's song in the ai
  ocean: A survey on hallucination in large language models}.

\bibitem[{Zhang et~al.(2023{\natexlab{b}})Zhang, Fang, Chen, Namazi-Rad, and
  Wang}]{zhang2023large}
Zihan Zhang, Meng Fang, Ling Chen, Mohammad-Reza Namazi-Rad, and Jun Wang.
  2023{\natexlab{b}}.
\newblock \href {http://arxiv.org/abs/2310.07343} {How do large language models
  capture the ever-changing world knowledge? a review of recent advances}.

\bibitem[{Zhao et~al.(2023{\natexlab{a}})Zhao, Li, Joty, Qin, and
  Bing}]{zhao2023verifyandedit}
Ruochen Zhao, Xingxuan Li, Shafiq Joty, Chengwei Qin, and Lidong Bing.
  2023{\natexlab{a}}.
\newblock \href {http://arxiv.org/abs/2305.03268} {Verify-and-edit: A
  knowledge-enhanced chain-of-thought framework}.

\bibitem[{Zhao et~al.(2023{\natexlab{b}})Zhao, Zhou, Li, Tang, Wang, Hou, Min,
  Zhang, Zhang, Dong, Du, Yang, Chen, Chen, Jiang, Ren, Li, Tang, Liu, Liu,
  Nie, and Wen}]{zhao2023survey}
Wayne~Xin Zhao, Kun Zhou, Junyi Li, Tianyi Tang, Xiaolei Wang, Yupeng Hou,
  Yingqian Min, Beichen Zhang, Junjie Zhang, Zican Dong, Yifan Du, Chen Yang,
  Yushuo Chen, Zhipeng Chen, Jinhao Jiang, Ruiyang Ren, Yifan Li, Xinyu Tang,
  Zikang Liu, Peiyu Liu, Jian-Yun Nie, and Ji-Rong Wen. 2023{\natexlab{b}}.
\newblock \href {http://arxiv.org/abs/2303.18223} {A survey of large language
  models}.

\bibitem[{Zheng et~al.(2023)Zheng, Li, Dong, Fan, Wu, Xu, and
  Chang}]{zheng2023can}
Ce~Zheng, Lei Li, Qingxiu Dong, Yuxuan Fan, Zhiyong Wu, Jingjing Xu, and Baobao
  Chang. 2023.
\newblock Can we edit factual knowledge by in-context learning?
\newblock \emph{arXiv preprint arXiv:2305.12740}.

\bibitem[{Zhong et~al.(2023)Zhong, Wu, Manning, Potts, and
  Chen}]{zhong2023mquake}
Zexuan Zhong, Zhengxuan Wu, Christopher~D Manning, Christopher Potts, and Danqi
  Chen. 2023.
\newblock Mquake: Assessing knowledge editing in language models via multi-hop
  questions.
\newblock \emph{arXiv preprint arXiv:2305.14795}.

\bibitem[{Zhou et~al.(2023)Zhou, Zhang, Poon, and
  Chen}]{zhou2023contextfaithful}
Wenxuan Zhou, Sheng Zhang, Hoifung Poon, and Muhao Chen. 2023.
\newblock \href {http://arxiv.org/abs/2303.11315} {Context-faithful prompting
  for large language models}.

\end{thebibliography}
